\documentclass[10pt,twocolumn,letterpaper]{article}

\usepackage{cvpr}              

\usepackage{graphicx}
\usepackage{amsmath}
\usepackage{amssymb}
\usepackage{booktabs}
\usepackage{color}
\usepackage{multicol}
\usepackage{caption}
\usepackage{mathtools}
\usepackage{cuted}

\usepackage{bm}
\usepackage[accsupp]{axessibility}

\usepackage[pagebackref,breaklinks,colorlinks]{hyperref}

\usepackage[capitalize]{cleveref}
\crefname{section}{Sec.}{Secs.}
\Crefname{section}{Section}{Sections}
\Crefname{table}{Table}{Tables}
\crefname{table}{Tab.}{Tabs.}
\newcommand{\printfnsymbol}[1]{%
        \textsuperscript{\@fnsymbol{#1}}%
}

\def\cvprPaperID{4531} 
\def\confName{CVPR}
\def\confYear{2023}

\newcommand{\ssecspace}{\vspace{-0.4em}}

\newcommand{\secspace}{\vspace{-0.0em}}

\begin{document}

\setlength{\abovedisplayskip}{0.2em}
\setlength{\belowdisplayskip}{0.2em}

\title{
\vspace{-1.0em}
SINE: Semantic-driven Image-based NeRF Editing \\with Prior-guided Editing Field
\vspace{-1.0em}
}

\author{
Chong Bao$^{1}$\footnotemark[1]
\quad
Yinda Zhang$^{2}$\footnotemark[1] \quad
Bangbang Yang$^{1}$\footnotemark[1]  \quad
Tianxing Fan$^{1}$\\
Zesong Yang$^{1}$ \quad
Hujun Bao$^{1}$ \quad
Guofeng Zhang$^{1}$\footnotemark[2] \quad
Zhaopeng Cui$^{1}$\footnotemark[2] \\
$^{1}$State Key Lab of CAD\&CG, Zhejiang University \quad
$^{2}$Google \\
{\url{https://zju3dv.github.io/sine/}}
}

\twocolumn[{%
\renewcommand\twocolumn[1][]{#1}%
\vspace{-3.6em}
\maketitle
\vspace{-3.3em}
\begin{center}
    \centering
    \includegraphics[width=1.0\linewidth, trim={0 0 0 0}, clip]{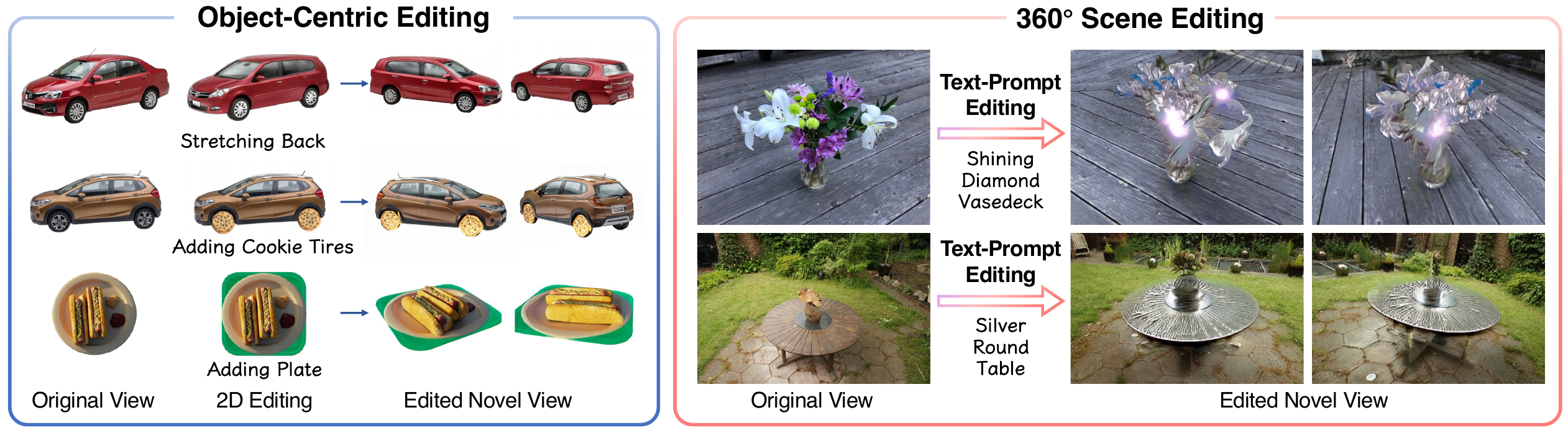}
    \captionof{figure}{
    We propose a novel semantic-driven image-based editing approach, which allows users to edit a photo-realistic NeRF with a single-view image or with text prompts, and renders edited novel views with vivid details and multi-view consistency.
    }
    \vspace{1.0em}
    \label{fig:teaser}
\end{center}%
}]
\renewcommand{\thefootnote}{\fnsymbol{footnote}}
\footnotetext[1]{Authors contributed equally.}
\footnotetext[2]{Corresponding authors.}

\begin{abstract}
Despite the great success in 2D editing using user-friendly tools, such as Photoshop, semantic strokes, or even text prompts, similar capabilities in 3D areas are still limited, either relying on 3D modeling skills or allowing editing within only a few categories.
In this paper, we present a novel semantic-driven NeRF editing approach, which enables users to edit a neural radiance field with a single image, and faithfully delivers edited novel views with high fidelity and multi-view consistency.
To achieve this goal, we propose a prior-guided editing field to encode fine-grained geometric and texture editing in 3D space, and develop a series of techniques to aid the editing process, including cyclic constraints with a proxy mesh to facilitate geometric supervision, a color compositing mechanism to stabilize semantic-driven texture editing, and a feature-cluster-based regularization to preserve the irrelevant content unchanged.
Extensive experiments and editing examples on both real-world and synthetic data demonstrate that our method achieves photo-realistic 3D editing using only a single edited image, pushing the bound of semantic-driven editing in 3D real-world scenes.
\end{abstract}
\vspace{-2.0em}

\section{Introduction}

Semantic-driven editing approaches, such as stroke-based scene editing~\cite{sdedit,editgan,wang2018high}, text-driven image synthesis and editing~\cite{diffusion,text2live,patashnik2021styleclip}, 
and attribute-based face editing~\cite{sun2022fenerf,conerf},
have greatly improved the ease of artistic creation.
However, despite the great success of 2D image editing and neural rendering techniques~\cite{nerf,dellaert2020neural}, similar editing abilities in the 3D area are still limited:
\textbf{(1)} they require laborious annotation such as image masks~\cite{object_nerf,conerf} and mesh vertices~\cite{neumesh,nerf_editing} to achieve the desired manipulation;
\textbf{(2)} they conduct global style transfer~\cite{arf,chiang2022stylizing,chen2022upstnerf,huang2022stylizednerf,fan2022unified} while ignoring the semantic meaning of each object part (\eg, windows and tires of a vehicle should be textured differently);
\textbf{(3)}
they can edit on categories by learning a textured 3D latent representation (\eg, 3D-aware GANs with faces and cars \etc)~\cite{pi-GAN,pix2nerf,eg3d,sun2022fenerf,gu2021stylenerf,sun2022ide,niemeyer2021giraffe,schwarz2020graf},
or at a coarse level~\cite{edit_nerf,clip_nerf} with 
basic color assignment or object-level disentanglement~\cite{kobayashi2022decomposing}, but struggle to conduct texture editing on objects with photo-realistic textures or out-of-distribution characteristics.

Based on this observation, we believe that, on the way toward
semantic-driven 3D editing, the following properties should be ensured.
First, the operation of editing should be effortless, \ie, users can edit 3D scenes on a single 2D image in convenient ways, 
\eg, using off-the-shelf tools such as GAN-based editing~\cite{editgan,stylegan}, text-driven editing~\cite{text2live,diffusion}, Photoshop, or even a downloaded Internet image without pixel-wise alignment, rather than steering 3D modeling software with specific knowledge~\cite{neumesh}, or repeatedly editing from multi-view images.
Second, the editing method should be applicable to real-world scenes or objects and preserve vivid appearances, which is beyond existing 3D-aware generative models~\cite{pi-GAN,eg3d} due to the limited categories and insufficient data diversity on real-world objects.

\begin{figure*}[!t]
    \centering
    \vspace{-1.7em}
    \includegraphics[width=0.97\linewidth, trim={0 0 0 0}, clip]{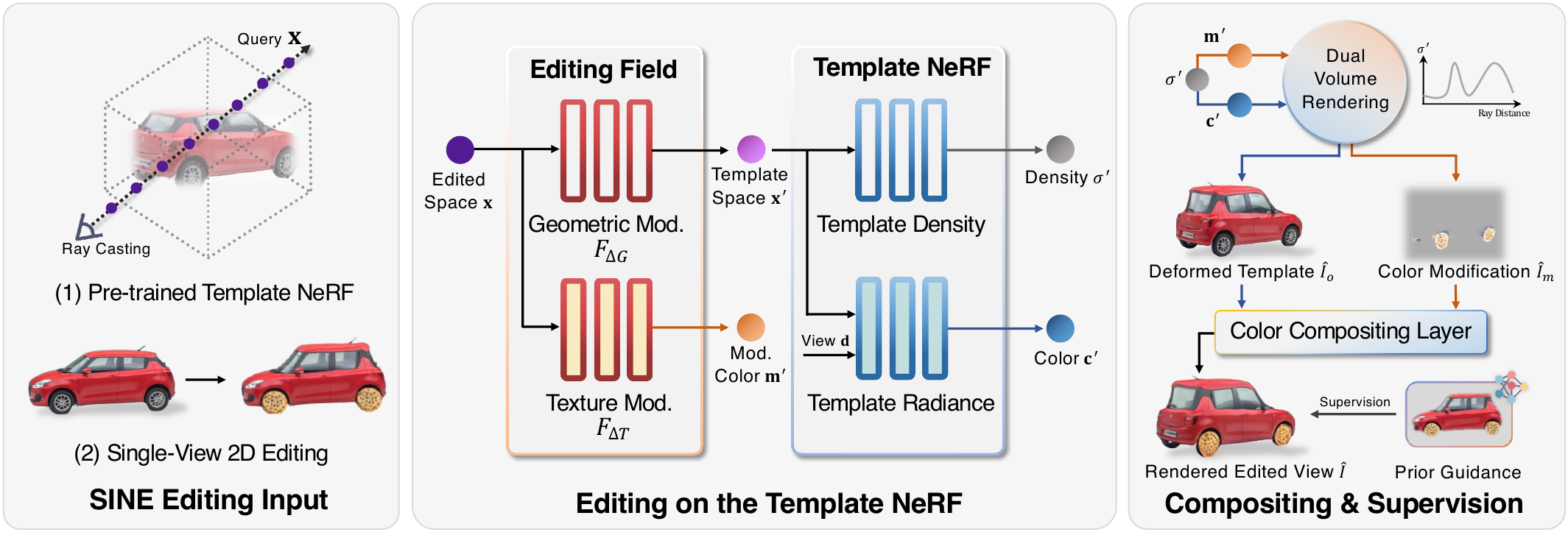}
    \caption{
    \textbf{Overview.}
    We encode geometric and texture changes over the original template NeRF with a prior-guided editing field, where the geometric modification field $F_{\Delta G}$ transformed the edited space query $\mathbf{x}$ into the template space $\mathbf{x}'$, and the texture modification field $F_{\Delta T}$ encodes modification colors $\mathbf{m}'$.
    Then, we render deformed template image $\hat{I}_{o}$ and color modification image $\hat{I}_{m}$ with all the queries, and use a color compositing layer to blend $\hat{I}_{o}$ and $\hat{I}_{m}$ into the edited view $\hat{I}$.
    \vspace{-0.5em}
    }
    \label{fig:framework}
\end{figure*}

To fulfill this goal, we propose a novel \textbf{S}emantic-driven \textbf{I}mage-based \textbf{E}diting approach for \textbf{N}eural radiance field in real-world scenes, named SINE.
Specifically, our method allows users to edit a neural radiance field with a
 single image, \ie, either by changing a rendered image using off-the-shelf image editing tools or providing an image for texture transferring (see Sec.~\ref{ssec:expr_tex_edit}),
and then delivers edited novel views with consistent semantic meaning.
Unlike previous works that directly fine-tune the existing NeRF model~\cite{clip_nerf,edit_nerf,kobayashi2022decomposing}, SINE learns a prior-guided editing field to encode geometric and texture changes over the original 3D scene (see Fig.~\ref{fig:framework}), thus enabling fine-grained editing ability.
By leveraging guidance from existing neural priors (shape prior models~\cite{dif} and Vision Transformer models~\cite{dino}, \etc), SINE can directly perform semantic-driven editing on photo-realistic scenes without pre-training a category-level latent space.
For example, in Fig.~\ref{fig:teaser}, users can stretch a car's back or change all four tires to cookies by only editing a single image, and can even cooperate with text-prompts editing~\cite{text2live} to modify a specific object of a scene with vivid appearances.

However, even when guided with neural priors, editing NeRF from a single image with multi-view consistency and accuracy is still challenging.
\textbf{(1)}
The generic NeRF does not necessarily provide an explicit surface or signed distance field, such that it cannot directly work with shape priors~\cite{dif}.
Therefore, we propose to use cyclic constraints with a proxy mesh to represent the edited NeRF's geometry, which facilitates guided editing using coarse shape prior.
\textbf{(2)}
Learning a coordinate-based 3D editing field using a single edited view is not sufficient to capture fine-grained details, and applying semantic supervision~\cite{dino,clip} directly to the editing field leads to sub-optimal convergence (see Sec.~\ref{ssec:expr_edit_ablation}).
To tackle these challenges, we propose a color compositing mechanism by first rendering the template NeRF color and modification color individually, and then deferred blending them to yield the edited view, which significantly improves semantic-driven texture editing.
\textbf{(3)}
Ideally, a user's editing should only affect the desired regions while maintaining other parts untouched.
However, in semantic-driven editing,
the prior losses require taking the full shape or image as input,
which leads to appearance or shape drifting at the undesired area.
To precisely control the editing while excluding irrelevant parts from being affected,
we generate feature clusters of the editing area using the ViT-based feature field~\cite{dino,kobayashi2022decomposing}, and use these clusters to distinguish whether a location is allowed to be edited or should remain unchanged.

In summary, the contributions of our paper are as follows.
\textbf{(1)}
We propose a novel semantic-driven image-based NeRF editing approach, called SINE, which allows users to edit a neural radiance field simply on just a single view of the rendering.
SINE leverages a prior-guided editing field to encode fine-grained geometry and texture changes over the given pre-trained NeRF,
thus delivering multi-view consistent edited views with high fidelity.
\textbf{(2)}
To achieve semantic editing functionality, we develop a series of techniques, 
including cyclic constraints with a proxy mesh for geometric editing,
the color compositing mechanism to enhance texture editing, and the feature-cluster-based regularization to control the affected editing area and maintain irrelevant parts unchanged.
\textbf{(3)}
Experiments and editing examples on both real-world/synthetic and object-centric/unbounded 360$^{\circ}$ scenes data demonstrate superior editing capabilities and quality with effortless operations.

\secspace
\vspace{-0.25em}

\section{Related Works}

\vspace{-0.25em}

\secspace

\noindent\textbf{Neural rendering with external priors.}
Neural rendering techniques aim at rendering novel views with high-quality~\cite{nerf} or controllable properties~\cite{park2021nerfies,conerf} by learning from 2D photo capture.
Recently, NeRF~\cite{nerf} achieves photo-realistic rendering with volume rendering and inspires many works, including surface reconstruction~\cite{neus,volsdf,li2022vox}, scene editing~\cite{guo2020object,object_nerf,wu2022objectsdf,yang2022neural,neural_outdoor_rerender} and generation~\cite{jain2022zero,poole2022dreamfusion}, inverse rendering~\cite{zhang2021nerfactor,boss2021neural}, SLAM~\cite{zhu2022nice, yang2022vox}, \etc.
For learning from few-shot images~\cite{jain2021putting} or 3D inpainting~\cite{mirzaei2022laterf}, NeRF's variants use hand-crafted losses~\cite{niemeyer2022regnerf} or large language-image models~\cite{jain2021putting,xu2022sinnerf} as external priors.
However, due to insufficient 3D supervision, such methods cannot reconstruct accurate geometry and only produce visually plausible results.
Besides, some works~\cite{mi2022im2nerf,li2022symmnerf,insafutdinov2022snes} use the symmetric assumption to reconstruct category-level objects (\eg, cars, chairs) but cannot generalize on complex scenes.

\noindent\textbf{Neural 2D \& 3D scene editing.}
With the development of neural networks,
semantic-driven 2D photo editing 
allows user editing in various friendly ways, such as controlling attribute of faces~\cite{he2019attgan,stylegan}, stroke-based editing~\cite{sdedit,editgan,wang2018high}, sketch-to-image generation~\cite{chen2018sketchygan,sangkloy2017scribbler}, image-to-image texture transferring~\cite{splice}, or text-driven image generation~\cite{diffusion} and editing~\cite{imagic}.
Nevertheless, in 3D scene editing, 
similar capabilities are still limited due to the high demand for multi-view consistency.
Existing approaches either rely on laborious annotation~\cite{conerf,nerf_editing,neumesh,object_nerf}, only support object deformation or translation~\cite{nerf_editing,kobayashi2022decomposing,feature_fusion_field,vora2021nesf}, or only perform global style transfer~\cite{arf,chiang2022stylizing,chen2022upstnerf,huang2022stylizednerf,fan2022unified} without strong semantic meaning.
Recently, 3D-aware GANs~\cite{pi-GAN,eg3d,gu2021stylenerf,sun2022ide,niemeyer2021giraffe,schwarz2020graf,jang2021codenerf} and semantic NeRF editing~\cite{edit_nerf,clip_nerf} learn a latent space of the category and enable editing via latent code control.
However, the quality and editing ability of these methods mainly depend on the dataset (\eg, human faces~\cite{sun2022fenerf,sun2022ide} or objects in ShapeNet~\cite{chang2015shapenet}), and they cannot generalize to objects with rich appearances or out-of-distribution features~\cite{text2live}.
In contrast, our method allows for semantic-driven editing directly on the given photo-realistic NeRF, and uses a prior-guided editing field to learn fine-grained editing from only a single image.

\secspace

\section{Method}

\secspace

We first formulate the goal of our semantic NeRF editing task as follows.
As illustrated in the left part of Fig.~\ref{fig:framework}, 
given a pre-trained NeRF of a photo-realistic scene (named template NeRF),
we aim at editing the template NeRF using only a single-view 2D image, and then produce novel views with consistent semantic meaning (see Sec.~\ref{ssec:expr_geo_edit} and Sec.~\ref{ssec:expr_tex_edit}).
Note that na\"ively fine-tuning on the edited single view cannot obtain satisfactory results due to the spatial ambiguity and lack of multi-view supervision (see Sec.~\ref{ssec:expr_neumesh}).
Therefore, 
we propose to use a novel prior-guided editing field to encode fine-grained changes (Sec.~\ref{ssec:framework}) in 3D space, which leverages geometry and texture priors to guide the learning of semantic-driven editing (Sec.~\ref{ssec:geo_prior} and Sec.~\ref{ssec:tex_prior}).
Besides, to precisely control the editing area while maintaining other parts unchanged, we design editing regularization with feature cluster-based semantic masking (Sec.~\ref{ssec:edit_reg}).

\ssecspace

\subsection{SINE Rendering Pipeline}
\label{ssec:framework}

\ssecspace

As illustrated in Fig.~\ref{fig:framework}, we use a dedicated editing field to encode geometry and texture changes over the pre-trained template NeRF.
The editing field consists of an implicit geometric modification field $F_{\Delta G}$ and a texture modification field $F_{\Delta T}$, where $F_{\Delta G}$ deforms the query points from the observed edited space to the original template space, 
as $\mathbf{x}' := F_{\Delta G}(\mathbf{x})$, 
and $F_{\Delta T}$ encodes the modification color $\mathbf{m}'$, as $\mathbf{m}' := F_{\Delta T}(\mathbf{x})$.
Specifically, for each sampled query point $\{\mathbf{x}_i|i=1,...,N\}$ along the ray $\bm{r}$ with view direction $\mathbf{d}$, we first obtain the deformed points $\mathbf{x}'$ (in template space) and modification color $\mathbf{m}'$, and feed $\mathbf{x}'$ and $\mathbf{d}$ to the template NeRF to obtain the density $\delta '$ and template colors $\mathbf{c}'$.
Then, we perform dual volume rendering both on edited fields and template NeRF following the quadrature rules~\cite{nerf,quadrature_rule}, which is defined as:
\begin{equation}
\begin{split}
        & \hat{C}_o(\bm{r}) = \sum_{i=1}^{N} T_i \alpha_i {\mathbf{c}}_i', \;\;\;   \hat{C}_e(\bm{r}) = \sum_{i=1}^{N} T_i \alpha_i {\mathbf{m}}_i', \\
        & T_i = \exp{\left(-\sum_{j=1}^{i-1}{\sigma'}_j \delta_j\right)},
\label{eq:rendering}
\end{split}
\end{equation}
where $\alpha_i = 1-\exp{(-{{\sigma}'}_i \delta_i)}$, and $\delta_i$ is the distance between adjacent samples along the ray. 
In this way, we obtain the deformed template image $\hat{I}_o$ from the template NeRF's pixel color $\hat{C}_o(\bm{r})$ and color modification image $\hat{I}_{m}$ from the modification color $\hat{C}_e(\bm{r})$.
Finally, we apply the color compositing layer (see Sec.~\ref{ssec:tex_prior}) to blend $\hat{I}_o$ and $\hat{I}_{m}$ into the resulting edited views $\hat{I}$.

\ssecspace

\subsection{Prior-Guided Geometric Editing}
\label{ssec:geo_prior}

\ssecspace

In this section, we explain how to learn $F_{\Delta G}(\mathbf{x})$ with the geometric prior.

\noindent\textbf{Shape prior constraint on the edited NeRF.}
We leverage geometric prior models, such as neural implicit shape representation~\cite{dif,park2019deepsdf} or depth prediction~\cite{bhat2021adabins}, to mitigate the ambiguity of geometric editing based on editing from a single perspective.
\textbf{(1)} For objects within a certain shape category (\eg, cars, airplanes), we use DIF~\cite{dif}, in which the implicit SDF field and the prior mesh $\hat{M}_P$ can be generated with the condition of an optimizable latent code $\hat{\mathbf{z}}$.
We force the edited NeRF's geometry $\hat{M}_{E}$ to be explainable by a pre-trained DIF model with the geometric prior loss:
\begin{equation}
\begin{split}
   \mathcal{L}_{gp} = \; & \underset{\hat{\mathbf{z}}}{\text{min}}
   \big(\sum_{\mathclap{\mathbf{p}'\in \hat{M}_E}} f_{\text{SDF}}(\hat{\mathbf{z}}, \mathbf{p}') +  \lambda ||\hat{\mathbf{z}}||_2^2 \big) \\
   &+ \sum_{\mathbf{p}'_i\in \hat{M}_{E}} \min_{\mathbf{p}_t \in \hat{M}_{P}} ||\mathbf{p}'_i-\mathbf{p}_t||^2_2 \\ 
   &+  \sum_{\mathbf{p}_i\in \hat{M}_{P}} \min_{\mathbf{p}'_t \in \hat{M}_E} ||\mathbf{p}_i-\mathbf{p}'_t||^2_2.
   \label{eq:geo_prior}
\end{split}
\end{equation}
The first term encourages the sampled surface points on the edited NeRF's geometry $\hat{M}_{E}$ to lie on the manifold of DIF's latent space with an SDF loss $f_{\text{SDF}}$ and the latent code regularization~\cite{dif}.
The last two terms are Chamfer constraints, which enforce the $\hat{M}_{E}$ close to the DIF's periodically updated prior mesh $\hat{M}_P$~\cite{lorensen1987marching} by minimizing the closest surface points.
\textbf{(2)} For objects without a category-level prior, we can build a finalized shape prior $\hat{M}_P$ beforehand.
Practically, we find 3D deforming vertices with 2D correspondence~\cite{jiang2021cotr} and monocular depth prediction~\cite{bhat2021adabins}, and use ARAP~\cite{as_rigid_as_possible} to deform the proxy triangle mesh $M_{\Theta}$ to $\hat{M}_P$.
Then, we can inherit the Chamfer loss term in Eq.~\eqref{eq:geo_prior} for prior-guided supervision.

\begin{figure}[!t]
    \centering
    \includegraphics[width=1.0\linewidth, trim={0 0 0 0}, clip]{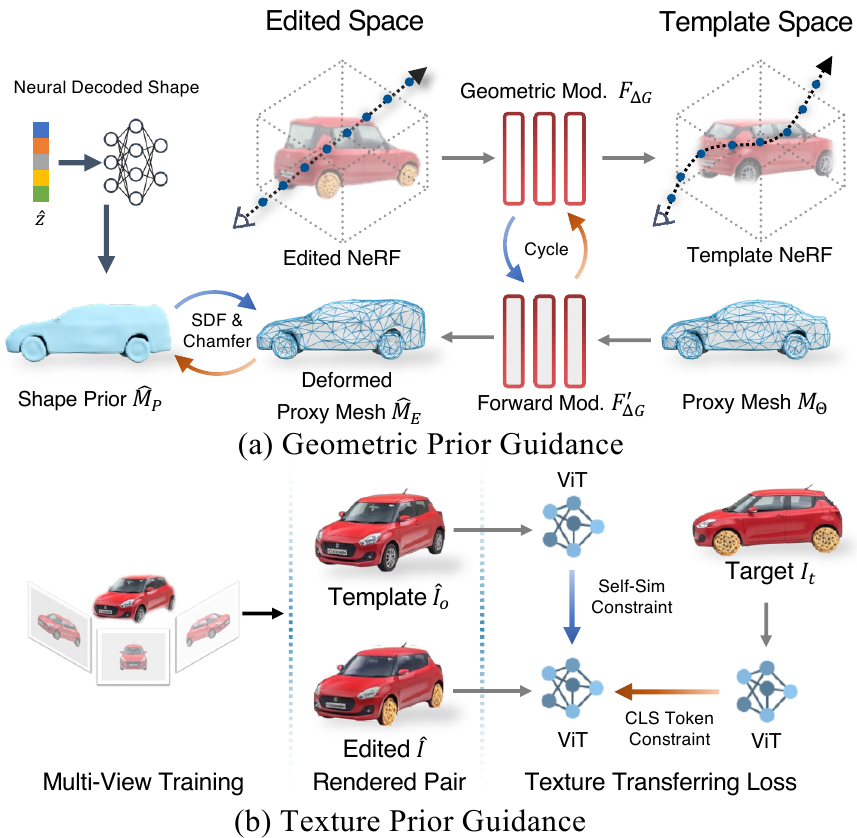}
    \caption{
    We leverage geometric~\cite{dif,bhat2021adabins} and texture~\cite{dino,splice} priors to guide the learning of semantic-driven NeRF editing. 
    }
    \label{fig:prior_guided_edit}
    \vspace{0.5em}
\end{figure}

\noindent\textbf{Representing edited NeRF's geometry as a deformed proxy mesh.}
The edited NeRF has no explicit surface definition or SDF field to directly apply the geometric prior loss (Eq.~\eqref{eq:geo_prior}).
Therefore, to obtain the edited mesh surface $\hat{M}_E$, as illustrated in Fig.~\ref{fig:prior_guided_edit} (a), we first fit the template NeRF geometry with a proxy mesh $M_{\Theta}$~\cite{lorensen1987marching,neus}, and then learn a forward modification field $F_{\Delta G}'$ to warp the template proxy mesh to the edited space.
$F_{\Delta G}'$ is an inverse of the editing field $F_{\Delta G}$, which maps from the template space to the query space~\cite{mihajlovic2021leap,neural_scene_flow}, as $\mathbf{x}:=F_{\Delta G}'(\mathbf{x}')$, and can be supervised using a cycle loss $\mathcal{L}_{\text{cyc}}$ (see the supplementary material for details). 
Note that the deformed mesh proxy might not reflect fine-grained details of the specific shape identity. It facilitates applying shape priors to the edited field and provides essential guidance during geometric editing.

\noindent\textbf{Learning geometric editing with users' 2D editing.}
The goal of geometric editing is to deform the given NeRF according to the edited target image while satisfying semantic properties.
To this end, apart from the geometric prior loss in Eq.~\eqref{eq:geo_prior}, we add the following geometric editing loss in two folds.
\textbf{(1)}
We encourage the edited NeRF to satisfy the user's edited image by directly supervising rendering colors and opacity on $N_r$ rays, which is defined as:
\begin{equation}
\begin{split}
    \mathcal{L}_{\text{gt}} = \frac{1}{|N_r|} \sum_{\bm{r}\in N_r}  ||\hat{C}(\bm{r})-C_t(\bm{r})||^2_2 +  \text{BCE}(\hat{O}(\bm{r}),O_e(\bm{r})).
\end{split}
\end{equation}
The first photometric loss term encourages the rendered color $\hat{C}(\bm{r})$ close to the edited target color $C_t(\bm{r})$.
The second silhouette loss term enforces the rendered opacity $\hat{O}(\bm{r})$ close to the edited object's silhouette $O_e(\bm{r})$ (derived from users' editing tools) by minimizing the binary cross-entropy loss, where $\hat{O}(\bm{r}) = \sum_{i=1}^{N} T_i \alpha_i$.
\textbf{(2)}
To obtain a spatially smooth deformation and mitigate overfitting to the mesh's surface points, inspired by previous works~\cite{park2021nerfies,park2021hypernerf,dif}, we also add deformation regularization as:
\begin{equation}
    \mathcal{L}_{\text{gr}}=\frac{1}{M} \sum_{i=1}^N ||\nabla F_{\Delta G}(\mathbf{p}_i)||_2 +  ||F_{\Delta G}(\mathbf{p}_i) -F_{\Delta G}(\mathbf{p}_i + \bm{\epsilon}) ||_1,
\end{equation}
where the first term penalizes the spatial gradient of the geometric editing, and the second term encourages the editing to be smooth under a mild 3D positional jitter $\bm{\epsilon}$.

The overall geometric editing loss is defined as:
\begin{equation}
    \mathcal{L}_{\text{geo}}=\lambda_{\text{gp}} \mathcal{L}_{\text{gp}}+\mathcal{L}_{\text{gt}} + \lambda_{\text{gr}} \mathcal{L}_{\text{gr}} + \lambda_{\text{cyc}} \mathcal{L}_{\text{cyc}} ,
\end{equation}
where we set $\lambda_{\text{gp}} = 0.03$, $\lambda_{\text{gr}} = 0.1$ and $\lambda_{\text{cyc}} = 10$.
Intuitively, the geometric editing loss $\mathcal{L}_{\text{geo}}$ jointly optimizes edited NeRF's geometry $\hat{M}_{E}$ and the latent shape code $\hat{\mathbf{z}}$ (for category-level objects) to best fit the user's 2D editing while maintaining shape prior's semantic properties (\ie, shape symmetry or physical conformity).

\ssecspace

\subsection{Prior-Guided Texture Editing}
\label{ssec:tex_prior}

\ssecspace

\noindent\textbf{Semantic texture prior supervision.}
In our task, users only conduct editing on a single image, but we hope to naturally propagate editing effects to multi-views with semantic meaning (see Fig.~\ref{fig:teaser}).
Therefore, we need to utilize semantic texture supervision that supports transferring the editing to the given NeRF across views, rather than using a pixel-aligned photometric loss.
Inspired by Tumanyan \etal~\cite{splice}, we use a pre-trained ViT model~\cite{dino} as the semantic texture prior, and apply the texture transferring loss in a multi-view manner as illustrated in Fig.~\ref{fig:prior_guided_edit} (b), which is defined as:
\begin{equation}
    \mathcal{L}_{\text{tex}} = || t_{\text{CLS}}(I_t) - t_{\text{CLS}}(\hat{I}) ||_2 + ||S(\hat{I}_o)-S(\hat{I})||_F,
\label{eq:tex_transfer}
\end{equation}
where $I_t$ is the user's edited image, $\hat{I}_o$ and $\hat{I}$ are the template image and edited image as introduced in Sec.~\ref{ssec:framework}.
$t_{\text{CLS}}(\cdot)$ and $S(\cdot)$ are the extracted deepest $\text{CLS}$ token and the structural self-similarity defined by Tumanyan \etal~\cite{splice}.
Essentially, this loss encourages $\hat{I}_o$ and $\hat{I}$ to share a similar spatial structure, and $\hat{I}_t$ and $\hat{I}$ to contain similar image cues.

\noindent\textbf{Decoupled rendering with color compositing layer.}
To achieve texture modification, a na\"ive approach is to directly add the modification color $\mathbf{m}'$ from the editing field to the template NeRF's radiance color $\mathbf{c}'$ during volume rendering.
However, we find it suffers from sub-optimal convergence when cooperating with texture transferring loss (see Sec.~\ref{ssec:expr_edit_ablation}), since NeRF struggles to learn the global-consistent appearance under the variational supervisory as shown in Fig.~\ref{fig:ablation_texture}(a).
To tackle this issue, we re-design the rendering pipeline in a decoupled manner.
As shown in Fig.~\ref{fig:framework},
we first render the deformed template image $\hat{I}_o$ with template NeRF and the color modification $\hat{I}_{m}$ with $F_{\Delta T}$, and then use a 2D CNN-based color compositing layer to deferred blend the modification $\hat{I}_{m}$ into the template image $\hat{I}_o$, which yields final edited view $\hat{I}$.
Intuitively, the coordinate-based editing field can encode fine-grained details from photometric constraints but cannot easily learn from coarse semantic supervision, while the proposed color compositing layer can reduce the difficulty by using easy-to-learn CNN layers before applying texture transferring loss.
Besides, it also learns view-dependent effects from the semantic prior, making the rendering results more realistic (\eg, the shining diamond effect in Fig.~\ref{fig:teaser}).

\ssecspace

\subsection{Editing Regularization}
\label{ssec:edit_reg}

\ssecspace

\noindent\textbf{Feature-cluster-based semantic masking.}
To precisely edit the desired region while preserving other content unchanged, inspired by previous works~\cite{vora2021nesf,kobayashi2022decomposing,feature_fusion_field}, we learn a distilled feature field with DINO-ViT~\cite{dino} to reconstruct scenes/objects with semantic features.
However, existing semantic field decomposing approaches~\cite{kobayashi2022decomposing,feature_fusion_field} are limited to the query-based similarity and require all the editing to be finalized on the 3D field, which is not compatible with our color compositing mechanism.
Therefore, we leverage users' editing silhouette $M_{e}$ to generate several feature clusters from the distilled feature map, and compute semantic masks $\hat{M}_{e}$ using the closest cosine similarity to cluster centers with a threshold, which will be served for image-based editing regularization.

\noindent\textbf{Regularization on geometric and texture editing.}
With the semantic masks that indicate the editing area, we can apply editing regularization to the geometric and texture editing, \ie, by enforcing the rendered pixels and the queries at the irrelevant part unchanged, which is defined as:
\begin{equation}
        \mathcal{L}_{\text{reg}} = \sum_{\mathclap{\mathbf{x} \in \hat{I} \setminus \hat{M}_{e}} }||F_{\Delta G}(\mathbf{x})||_1 + \sum_{\mathclap{\bm{r} \in \hat{I} \setminus \hat{M}_{e}}} ||\hat{C}(\mathbf{r}) - \hat{C}_o(\mathbf{r})||_2^2,
\label{eq:edit_reg}
\end{equation}
where the sampled points $\mathbf{x}$ and rays $\bm{r}$ are both from the background area of the computed semantic masks $\hat{M}_{e}$.

\begin{figure}[!t]
    \centering
    \includegraphics[width=0.97\linewidth, trim={0 0 0 0}, clip]{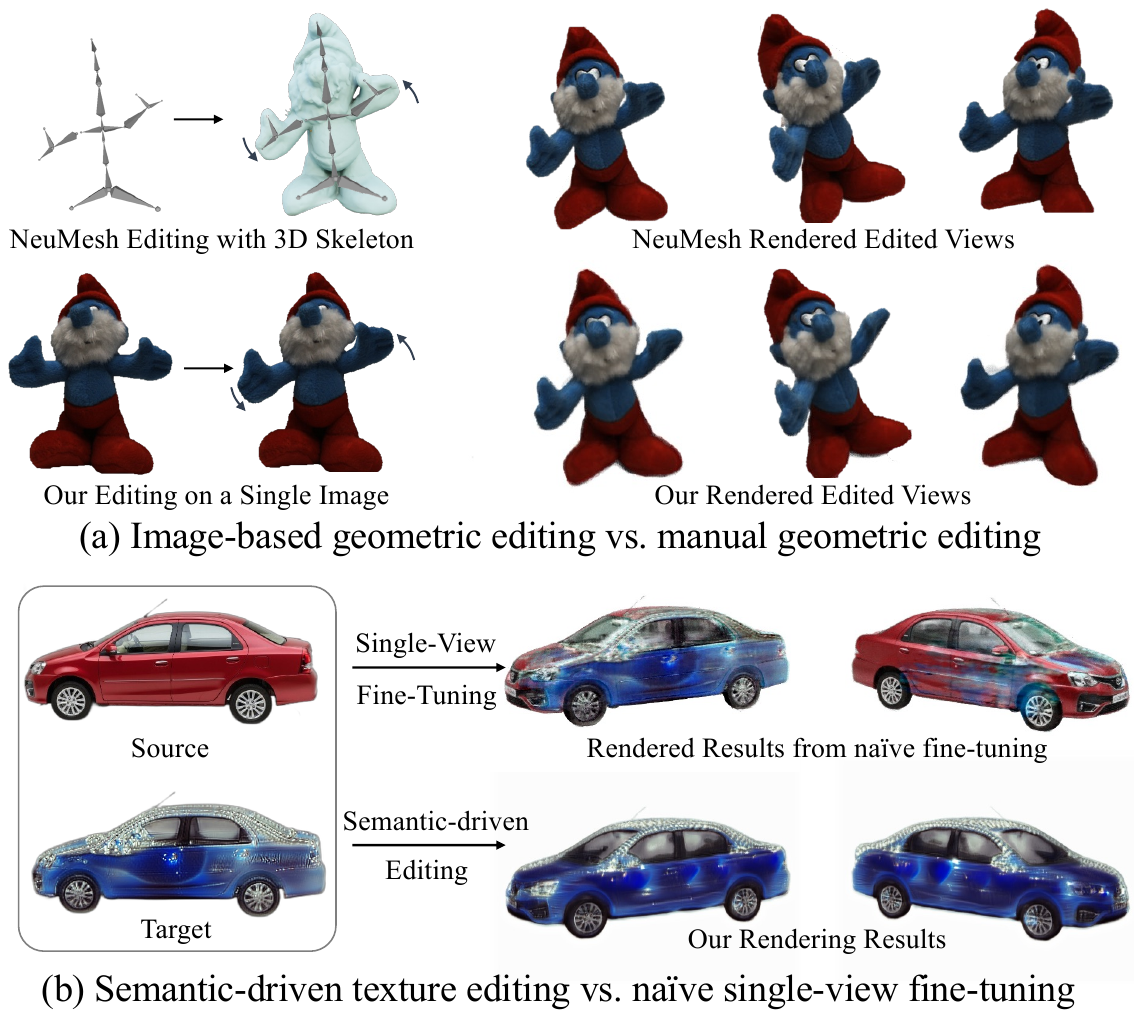}
    \caption{
    We show the difference between our semantic-driven image-based NeRF editing and manual NeRF editing~\cite{neumesh}.
    }
    \label{fig:edit_compare_neumesh}
    \vspace{0.3em}
\end{figure}

\begin{figure*}[!t]
    \centering
    \includegraphics[width=1.0\linewidth, trim={0 0 0 0}, clip]{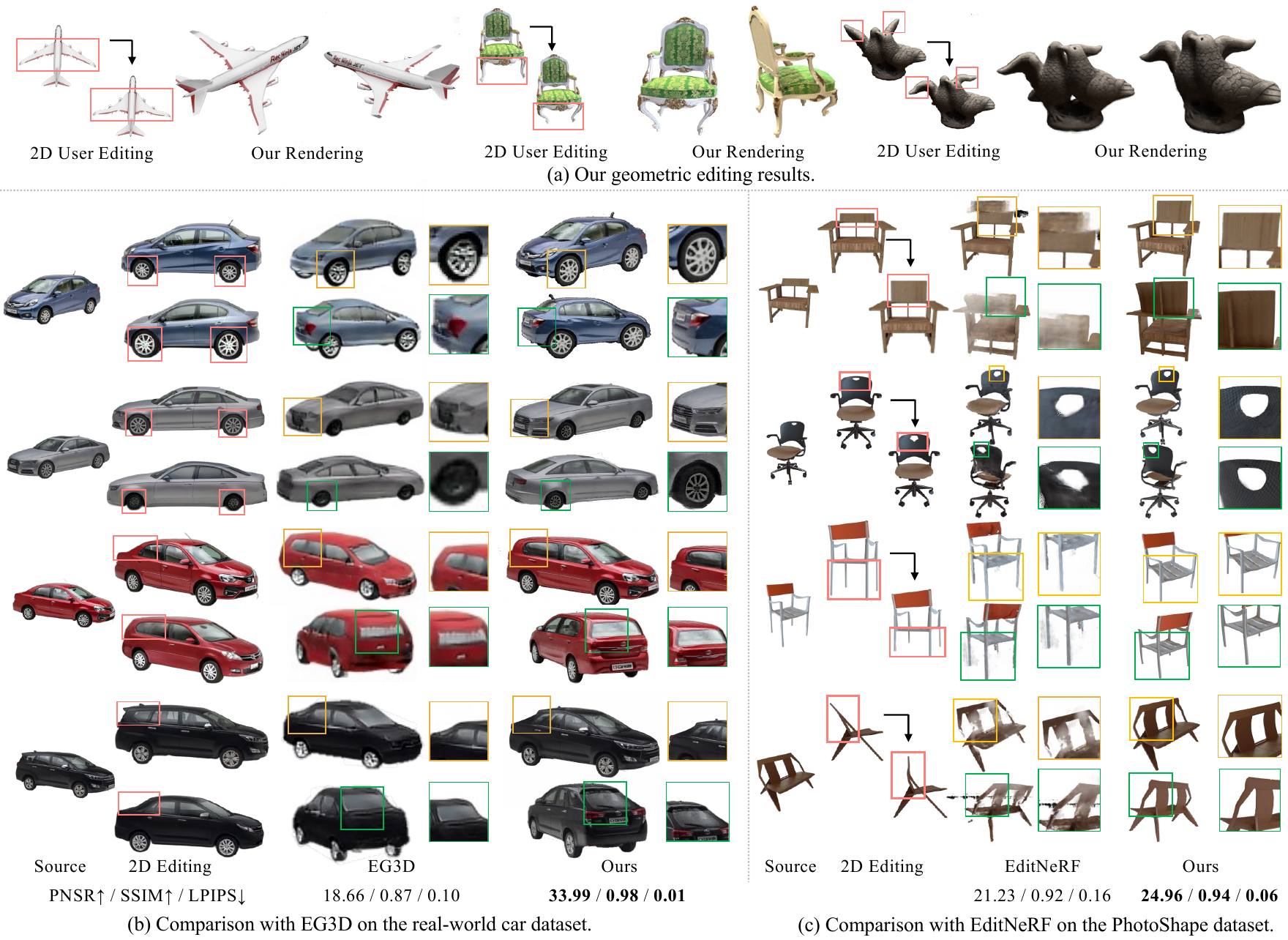}
    \caption{
    We compare the geometric editing with EG3D~\cite{eg3d} and EditNeRF~\cite{edit_nerf} on the real-world cars~\cite{CarWale} and PhotoShape~\cite{photoshape2018}.
    }
    \label{fig:geo_edit}
\end{figure*}

\secspace

\section{Experiments}

\ssecspace

\subsection{Datasets}
\label{ssec:expr_data}

\ssecspace

We evaluate SINE on both real-world/synthetic and object/scene datasets, including real-world car dataset~\cite{CarWale}, PhotoShape datasets (synthetic chairs)~\cite{photoshape2018}, ``pinecone'', ``vasedeck'', and ``garden'' from NeRF real-world 360$^{\circ}$ scenes~\cite{nerf,barron2022mipnerf360}, ``chairs'' and ``hotdog'' from NeRF photo-realistic synthetic data~\cite{nerf}, bird status from DTU~\cite{dtu} dataset, and ``airplane'' from BlenderSwap~\cite{blenderswap}.
Please refer to the supplementary materialfor more details.

\begin{figure*}[!t]
    \centering
    \includegraphics[width=1.0\linewidth, trim={0 0 0 0}, clip]{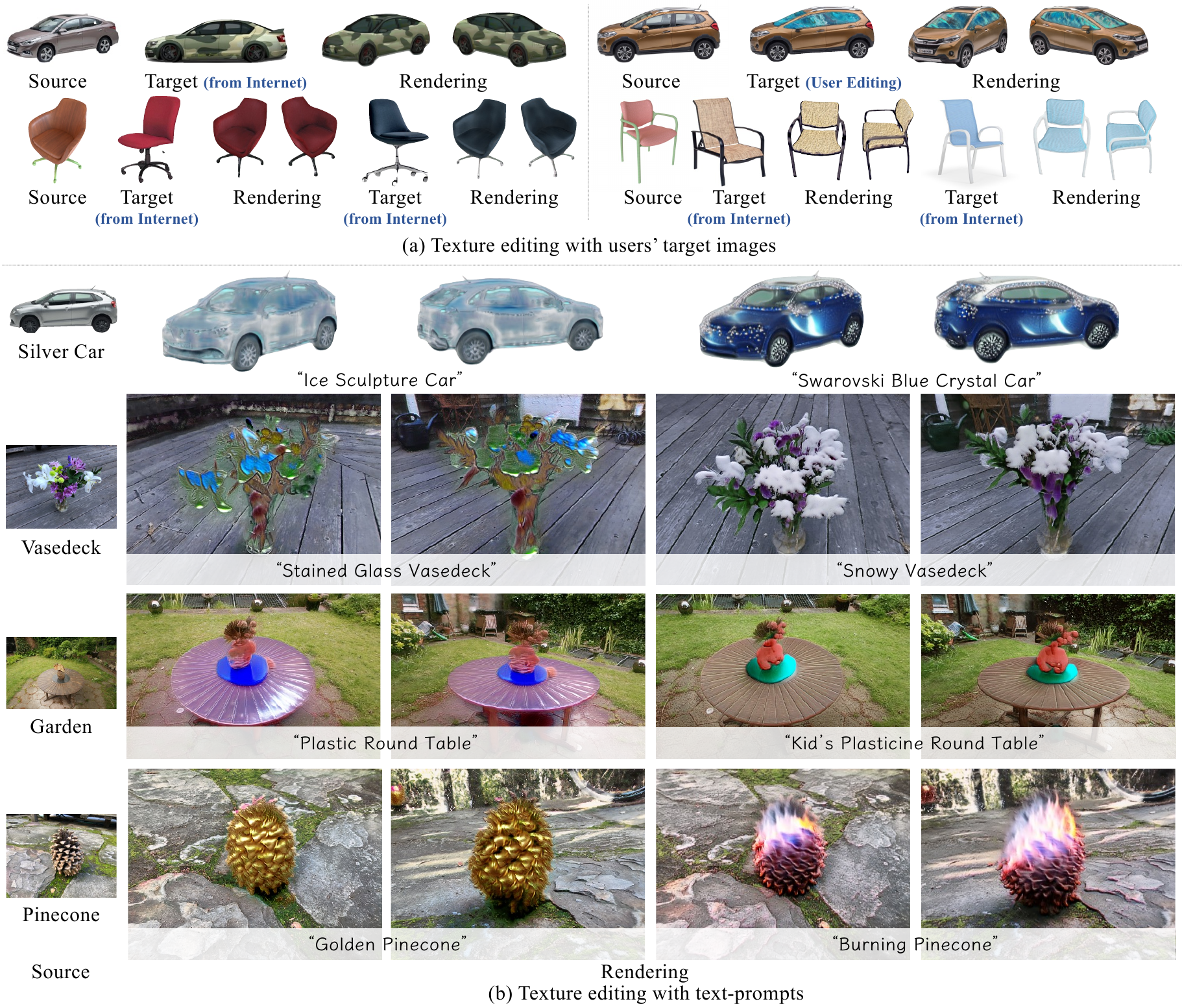}
    \caption{
    We show our texture editing results when given users' target images and cooperating with text-prompt-based editing methods~\cite{text2live}.
}
    \label{fig:our_tex_edit}
\end{figure*}

\ssecspace

\subsection{Semantic-driven vs. Manual Editing}
\label{ssec:expr_neumesh}

\ssecspace

We first clarify the difference between our semantic-driven NeRF editing and manual NeRF editing (\eg, NeuMesh~\cite{neumesh}, NeRF-Editing~\cite{nerf_editing}).
As illustrated in Fig.~\ref{fig:edit_compare_neumesh} (a), our method provides much more effortless ways than manual approaches.
For example, they require 3D modeling skills to bind the skeleton of the mesh using Blender~\cite{neumesh}, or drive models~\cite{nerf_editing} with Mixamo poses~\cite{Mixamo}, while our method can easily achieve similar geometric editing with only a single-view image.
Besides, na\"ively fine-tuning NeRF on a single-view with a pixel-aligned photometric loss like NeuMesh~\cite{neumesh} would only modify visible regions, which leads to inconsistent novel view rendering (\eg, in Fig.~\ref{fig:edit_compare_neumesh} (b), the car edited by single-view fine-tuning would expose unpainted red part).
On the contrary, our method leverages semantic priors~\cite{dino} to naturally edit objects with multi-view consistency, which does not require pixel-wise alignment and enables texture transferring between objects with different shapes (see cars and chairs in Fig.~\ref{fig:our_tex_edit} (a)).

\ssecspace

\subsection{Semantic-driven Geometric Editing}
\label{ssec:expr_geo_edit}

\ssecspace

We first show our geometric editing results in Fig.~\ref{fig:geo_edit} (a), where the objects can be faithfully deformed according to users' 2D editing (\eg, the airplane with warped wings~\cite{blenderswap}, green chair with bent legs~\cite{nerf} and deformed bird status~\cite{dtu}).
For the usage of geometry prior, we use a pre-trained DIF~\cite{dif} model for cars~\cite{CarWale}, chairs~\cite{photoshape2018} and planes~\cite{blenderswap}, and use ARAP-based shape priors for general objects without a category-level prior (\ie, toy in Fig.~\ref{fig:edit_compare_neumesh} (a), green
chair with unusual shape and status in Fig.~\ref{fig:geo_edit} (a), \etc).
Then, we compare our method with EG3D~\cite{eg3d}, a 3D-aware generative model that learns a latent representation of category-level objects, and EditNeRF~\cite{edit_nerf}, a NeRF-variant that supports object editing with single-view user interaction.
In addition to editing comparisons, we also used PSNR, SSIM, and LPIPS~\cite{nerf} to measure the edited rendering quality on synthetic cars and chairs.
For the geometric editing on EG3D, we first conduct 3D GAN inversion to obtain the style code with multi-view images
(same input as ours), and then fine-tune the code on the target images.
As shown in Fig.~\ref{fig:geo_edit} (b), we conduct different editing operations on four cars from CalWare 360$^{\circ}$ datasets with DIF shape prior~\cite{dif}, \ie, enlarging/shrinking tires/back.
Due to the difficulty of learning a latent textured 3D representation and the limitation of data diversity, 3D-aware generative models like EG3D cannot produce rendering results with fine-grained details, which also results in lower evaluation metrics.
For the Photoshape~\cite{photoshape2018}, EditNeRF~\cite{edit_nerf} does not provide edited GT images,  so we regenerate all testing cases using Blender, which is more challenging than the original ones.
Then, we evaluate EditNeRF~\cite{edit_nerf} by fine-tuning the pre-trained models on specific chairs from the PhotoShape dataset.
As shown in Fig.~\ref{fig:geo_edit} (c), EditNeRF produces more blurry rendering results than ours, and cannot achieve satisfactory results with single-view editing (\eg, multi-view inconsistent chair back in the first row, and unmodified or blurry shapes in the third and fourth rows).
By contrast, our method consistently delivers high-fidelity rendering results and achieves reliable editing capability by leveraging geometric priors~\cite{dif,bhat2021adabins}.
This demonstrates that, for semantic geometric NeRF editing, learning a prior-guided editing field like ours can maintain better visual quality and achieve greater generalization ability than pre-training a textured 3D generative model or latent model.

\begin{figure*}[!t]
    \centering
    \includegraphics[width=1.0\linewidth, trim={0 0 0 0}, clip]{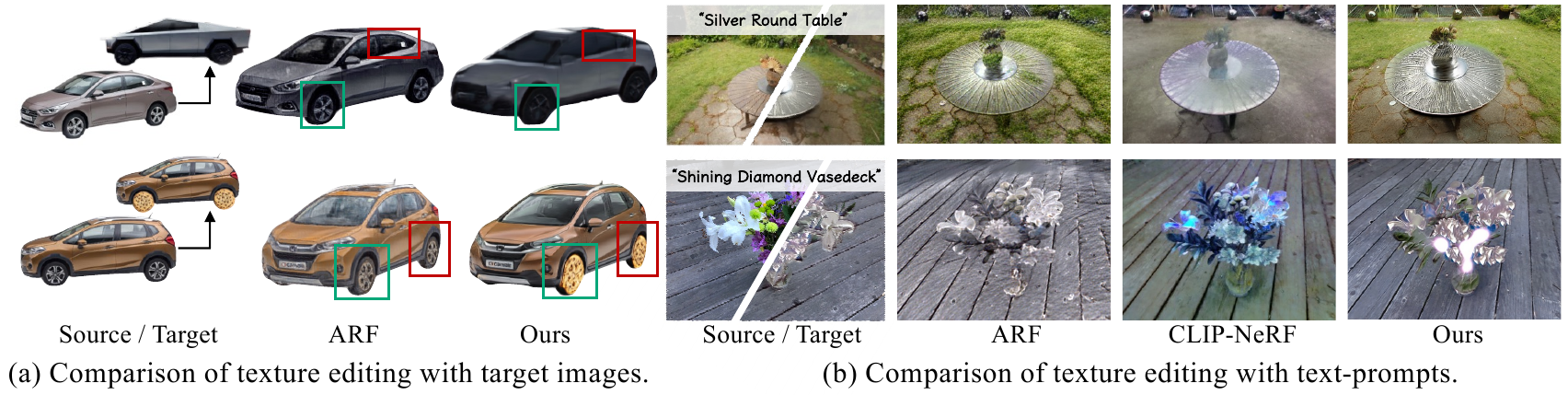}
    \caption{
    We compare our texture editing with ARF~\cite{arf} and CLIP-NeRF~\cite{clip_nerf} on the real-world cars~\cite{CarWale} and 360$^{\circ}$~\cite{nerf,barron2022mipnerf360} scene dataset.
}
    \label{fig:compare_tex_edit}
\end{figure*}

\begin{figure*}[!t]
    \centering
    \includegraphics[width=1.0\linewidth, trim={0 0 0 0}, clip]{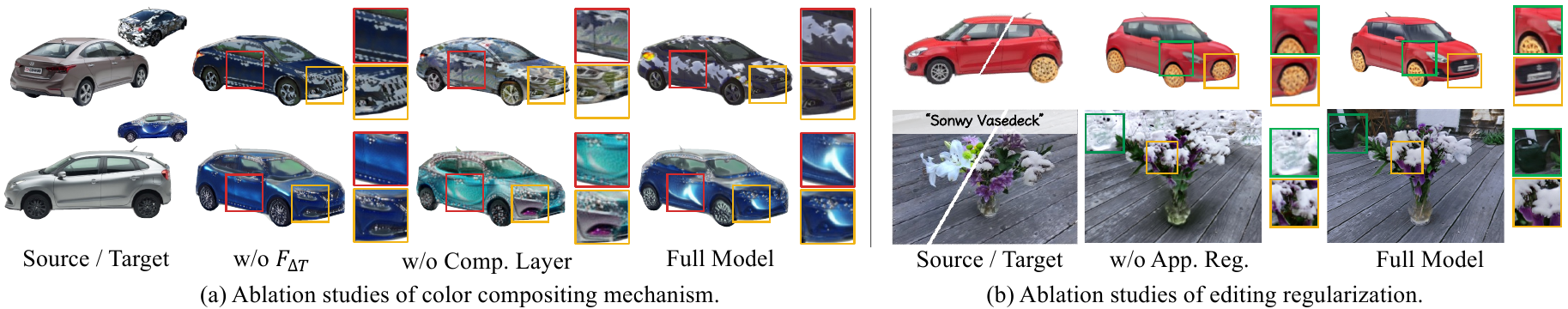}
    \caption{
    We analyze the effectiveness of the color compositing mechanism and editing regularization in texture editing. 
    }
    \label{fig:ablation_texture}
\end{figure*}

\begin{figure}[!t]
    \centering
    \vspace{0em}
    \includegraphics[width=1.0\linewidth, trim={0 0 0 0}, clip]{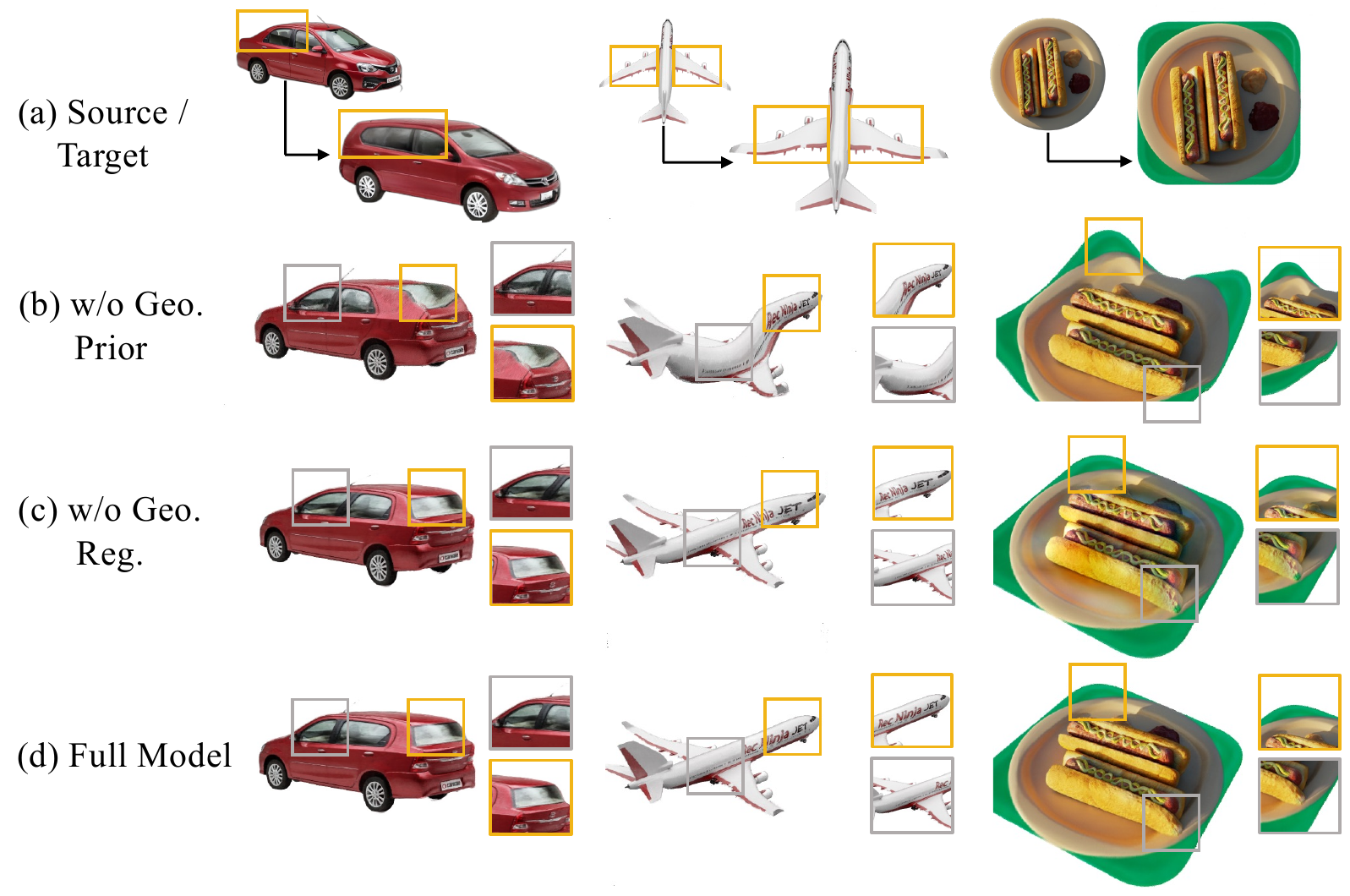}
    \caption{
    We inspect the efficacy of the geometric prior constraint and editing regularization in geometric editing.
    }
    \label{fig:ablation_geometry}
    \vspace{0.5em}
\end{figure}

\ssecspace

\subsection{Semantic-driven Texture Editing}
\label{ssec:expr_tex_edit}

\ssecspace

We evaluate our semantic texture editing ability on both objects (cars from CalWare 360$^{\circ}$, chairs from PhotoShape~\cite{photoshape2018}) and unbounded 360$^{\circ}$ scenes~\cite{nerf,barron2022mipnerf360}.
Since our method only requires a single image as editing input, we exhibit several editing functionalities as shown in Fig.~\ref{fig:our_tex_edit}.
Users can edit by assigning new textures on the car using Photoshop (adding sea wave windows in Fig.~\ref{fig:our_tex_edit} (a)), using a downloaded Internet image with different shapes as a reference (transferring textures of cars and chairs in Fig.~\ref{fig:our_tex_edit} (a)).
Moreover, we cooperate SINE with off-the-shelf text-prompts editing methods~\cite{text2live} by using a single text-edited image as the target, which enables to change the object's appearance in the 360$^{\circ}$ scene with vivid effects (\eg, shiny plastic round table or burning pinecone in Fig.\ref{fig:our_tex_edit} (b)) while preserving background unchanged.
It is noteworthy that our method does not pre-train a latent model within a specific category like cars or chairs, yet still transfers texture between objects with correct semantic meaning, \eg, the texture styles of chair legs and cloths in the edited views are precisely matched to the target images in Fig.~\ref{fig:our_tex_edit} (a).
Besides, we also compare our methods with ARF~\cite{arf}, a NeRF stylization method that also takes a single reference image as input, and CLIP-NeRF~\cite{clip_nerf}, which supports text-driven NeRF editing using the large language model~\cite{clip}.
As demonstrated in Fig.~\ref{fig:compare_tex_edit}, ARF globally changes appearance colors to the given target images but fails to produce fine-grained details (\eg, cookie tires in Fig.~\ref{fig:compare_tex_edit} (a)).
For CLIP-NeRF, since it directly fine-tunes NeRF's color layers, the results only show color/hue adjustment on the original scene (\eg, in Fig.~\ref{fig:compare_tex_edit}, the round table turns gray instead of a realistic silver texture, the vasedeck turns blue instead of a shining diamond).
Thanks to the prior-guided editing field, our method learns more fine-grained editing details than the others, and achieves texture editing with consistent semantic meaning to the given target images (\eg, similar appearance to the Tesla's cybertruck in Fig.~\ref{fig:compare_tex_edit} (a)), and delivers rich appearance details and vivid effects (\eg, silver texture and shining diamond effects in Fig.~\ref{fig:compare_tex_edit} (b)).

\noindent\textbf{User study.}
We also perform user studies to compare our texture editing (including target-image-based editing and text-prompts editing as Fig.~\ref{fig:compare_tex_edit}) with ARF~\cite{arf} and CLIP-NeRF~\cite{clip_nerf} on 43 cases with 30 users.
The results show that users prefer our methods (89.5\% / 83.3\%) to ARF (10.4\% / 7.4\%) and CLIP-NeRF (9.3\%).
Please refer to the supplementary material for more details.

\ssecspace

\subsection{Ablation Studies}
\label{ssec:expr_edit_ablation}

\ssecspace

\noindent\textbf{Geometric prior constraints.}
We first analyze the effectiveness of geometric prior constraints (Sec.~\ref{ssec:geo_prior}) by ablating geometric prior loss (Eq.~\eqref{eq:geo_prior}) in Fig.~\ref{fig:ablation_geometry} (\ie, deforming cars and airplanes with DIF shape prior, and adding plates with general shape prior).
As shown in Fig.~\ref{fig:ablation_geometry} (b), when learning without geometric prior constraints, the object will be distorted when rendered from other views (\eg, collapsed car back, twisted airplanes, and warped hotdog plates).
By applying geometric prior constraints, we successfully mitigate the geometric ambiguity for single-image-based editing and produce plausible rendering results from novel views.

\noindent\textbf{Color compositing mechanism.}
We then inspect the efficacy of the color compositing mechanism (Sec.~\ref{ssec:tex_prior}) by disabling the texture modification field $F_{\Delta T}$ and the color compositing layer in turn.
As demonstrated in Fig.~\ref{fig:ablation_texture} (a), when learning texture editing without $F_{\Delta T}$, the rendered edited object can show a similar global appearance to the target, but lose vivid local patterns (\eg, gray and white grains and blue shininess).
When ablating the color compositing layers, the editing effect might not be properly applied to every part of the object (\eg, the uncovered gray part of the car's front).
When all the compositing mechanism is enabled, we successfully learn NeRF editing with fine-grained local patterns and globally similar appearance.

\noindent\textbf{Editing regularization.}
We finally evaluate the editing regularization (Sec.~\ref{ssec:edit_reg}) in geometric and texture editing by ablating regularization loss (Eq.~\eqref{eq:edit_reg}).
As shown in Fig.~\ref{fig:ablation_geometry} (c) and Fig.~\ref{fig:ablation_texture} (b), when learning editing without regularization, the irrelevant part would be inevitably changed (\eg, bent car's front and airplane's head in Fig.~\ref{fig:ablation_geometry} (c), a spurious cookie at the car's front and snowy background in Fig.~\ref{fig:ablation_texture} (b)).
By adding editing regularization, we can modify the user-desired objects precisely while preserving other content unchanged.

Please refer to the supplementary material for more experiments (\eg, ablation on more loss terms, visualization of color composition layer, discussion with external supervision, \etc).

\section{Conclusion}

\secspace

We have proposed a novel semantic-driven NeRF editing approach, which supports editing a photo-realistic template NeRF with a single user-edited image, and deliver edited novel views with high-fidelity and multi-view consistency.
As limitation, our approach does not support editing with topology changes, which can be future work.
Besides, our method assumes users' editing to be semantically meaningful, so we cannot use target images with meaningless random paintings.

\noindent\textbf{Acknowledgment.} This work was partially supported by NSF of China~(No. 62102356).

\clearpage

{\small
\bibliographystyle{ieee_fullname}
\bibliography{main}

\begin{thebibliography}{10}\itemsep=-1pt

\bibitem{text2live}
Omer Bar-Tal, Dolev Ofri-Amar, Rafail Fridman, Yoni Kasten, and Tali Dekel.
\newblock Text2live: Text-driven layered image and video editing.
\newblock {\em arXiv preprint arXiv:2204.02491}, 2022.

\bibitem{barron2022mipnerf360}
Jonathan~T. Barron, Ben Mildenhall, Dor Verbin, Pratul~P. Srinivasan, and Peter
  Hedman.
\newblock Mip-nerf 360: Unbounded anti-aliased neural radiance fields.
\newblock {\em In Proceedings of the IEEE/CVF Conference on Computer Vision and
  Pattern Recognition (CVPR)}, 2022.

\bibitem{bhat2021adabins}
Shariq~Farooq Bhat, Ibraheem Alhashim, and Peter Wonka.
\newblock Adabins: Depth estimation using adaptive bins.
\newblock In {\em Proceedings of the IEEE/CVF Conference on Computer Vision and
  Pattern Recognition}, pages 4009--4018, 2021.

\bibitem{neural_outdoor_rerender}
{Boming Zhao and Bangbang Yang}, Zhenyang Li, Zuoyue Li, Guofeng Zhang, Jiashu
  Zhao, Dawei Yin, Zhaopeng Cui, and Hujun Bao.
\newblock Factorized and controllable neural re-rendering of outdoor scene for
  photo extrapolation.
\newblock In {\em Proceedings of the 30th ACM International Conference on
  Multimedia}, 2022.

\bibitem{boss2021neural}
Mark Boss, Varun Jampani, Raphael Braun, Ce Liu, Jonathan Barron, and Hendrik
  Lensch.
\newblock Neural-pil: Neural pre-integrated lighting for reflectance
  decomposition.
\newblock {\em Advances in Neural Information Processing Systems},
  34:10691--10704, 2021.

\bibitem{pix2nerf}
Shengqu Cai, Anton Obukhov, Dengxin Dai, and Luc Van~Gool.
\newblock Pix2nerf: Unsupervised conditional p-gan for single image to neural
  radiance fields translation.
\newblock In {\em Proceedings of the IEEE/CVF Conference on Computer Vision and
  Pattern Recognition (CVPR)}, pages 3981--3990, June 2022.

\bibitem{dino}
Mathilde Caron, Hugo Touvron, Ishan Misra, Herv{\'e} J{\'e}gou, Julien Mairal,
  Piotr Bojanowski, and Armand Joulin.
\newblock Emerging properties in self-supervised vision transformers.
\newblock In {\em Proceedings of the IEEE/CVF International Conference on
  Computer Vision}, pages 9650--9660, 2021.

\bibitem{pi-GAN}
Eric Chan, Marco Monteiro, Petr Kellnhofer, Jiajun Wu, and Gordon Wetzstein.
\newblock pi-gan: Periodic implicit generative adversarial networks for
  3d-aware image synthesis.
\newblock In {\em Proceedings of the IEEE/CVF conference on computer vision and
  pattern recognition (CVPR)}, 2021.

\bibitem{eg3d}
Eric~R. Chan, Connor~Z. Lin, Matthew~A. Chan, Koki Nagano, Boxiao Pan,
  Shalini~De Mello, Orazio Gallo, Leonidas Guibas, Jonathan Tremblay, Sameh
  Khamis, Tero Karras, and Gordon Wetzstein.
\newblock Efficient geometry-aware {3D} generative adversarial networks.
\newblock In {\em Proceedings of the IEEE/CVF Conference on Computer Vision and
  Pattern Recognition (CVPR)}, 2022.

\bibitem{chang2015shapenet}
Angel~X Chang, Thomas Funkhouser, Leonidas Guibas, Pat Hanrahan, Qixing Huang,
  Zimo Li, Silvio Savarese, Manolis Savva, Shuran Song, Hao Su, et~al.
\newblock Shapenet: An information-rich 3d model repository.
\newblock {\em arXiv preprint arXiv:1512.03012}, 2015.

\bibitem{chen2018sketchygan}
Wengling Chen and James Hays.
\newblock Sketchygan: Towards diverse and realistic sketch to image synthesis.
\newblock In {\em Proceedings of the IEEE Conference on Computer Vision and
  Pattern Recognition}, pages 9416--9425, 2018.

\bibitem{chen2022upstnerf}
Yaosen Chen, Qi Yuan, Zhiqiang Li, Yuegen Liu, Wei Wang, Chaoping Xie, Xuming
  Wen, and Qien Yu.
\newblock {UPST-NeRF}: Universal photorealistic style transfer of neural
  radiance fields for 3d scene.
\newblock In {\em arXiv preprint arXiv:2208.07059}, 2022.

\bibitem{chiang2022stylizing}
Pei-Ze Chiang, Meng-Shiun Tsai, Hung-Yu Tseng, Wei-Sheng Lai, and Wei-Chen
  Chiu.
\newblock Stylizing 3d scene via implicit representation and hypernetwork.
\newblock In {\em Proceedings of the IEEE/CVF Winter Conference on Applications
  of Computer Vision}, pages 1475--1484, 2022.

\bibitem{dellaert2020neural}
Frank Dellaert and Lin Yen-Chen.
\newblock Neural volume rendering: Nerf and beyond.
\newblock {\em arXiv preprint arXiv:2101.05204}, 2020.

\bibitem{dif}
Yu Deng, Jiaolong Yang, and Xin Tong.
\newblock Deformed implicit field: Modeling 3d shapes with learned dense
  correspondence.
\newblock In {\em Proceedings of the IEEE/CVF Conference on Computer Vision and
  Pattern Recognition}, pages 10286--10296, 2021.

\bibitem{fan2022unified}
Zhiwen Fan, Yifan Jiang, Peihao Wang, Xinyu Gong, Dejia Xu, and Zhangyang Wang.
\newblock Unified implicit neural stylization.
\newblock {\em arXiv preprint arXiv:2204.01943}, 2022.

\bibitem{removebg}
Kaleido~AI GmbH.
\newblock removebg.bg.
\newblock \url{https://www.remove.bg/}, 2022.
\newblock {A}ccessed: 2022-11-10.

\bibitem{gu2021stylenerf}
Jiatao Gu, Lingjie Liu, Peng Wang, and Christian Theobalt.
\newblock Stylenerf: A style-based 3d-aware generator for high-resolution image
  synthesis.
\newblock {\em arXiv preprint arXiv:2110.08985}, 2021.

\bibitem{guo2020object}
Michelle Guo, Alireza Fathi, Jiajun Wu, and Thomas Funkhouser.
\newblock Object-centric neural scene rendering.
\newblock {\em arXiv preprint arXiv:2012.08503}, 2020.

\bibitem{he2019attgan}
Zhenliang He, Wangmeng Zuo, Meina Kan, Shiguang Shan, and Xilin Chen.
\newblock Attgan: Facial attribute editing by only changing what you want.
\newblock {\em IEEE transactions on image processing}, 28(11):5464--5478, 2019.

\bibitem{huang2022stylizednerf}
Yi-Hua Huang, Yue He, Yu-Jie Yuan, Yu-Kun Lai, and Lin Gao.
\newblock Stylizednerf: consistent 3d scene stylization as stylized nerf via
  2d-3d mutual learning.
\newblock In {\em Proceedings of the IEEE/CVF Conference on Computer Vision and
  Pattern Recognition}, pages 18342--18352, 2022.

\bibitem{insafutdinov2022snes}
Eldar Insafutdinov, Dylan Campbell, Jo{\~a}o~F Henriques, and Andrea Vedaldi.
\newblock Snes: Learning probably symmetric neural surfaces from incomplete
  data.
\newblock {\em arXiv preprint arXiv:2206.06340}, 2022.

\bibitem{jain2022zero}
Ajay Jain, Ben Mildenhall, Jonathan~T Barron, Pieter Abbeel, and Ben Poole.
\newblock Zero-shot text-guided object generation with dream fields.
\newblock In {\em Proceedings of the IEEE/CVF Conference on Computer Vision and
  Pattern Recognition}, pages 867--876, 2022.

\bibitem{jain2021putting}
Ajay Jain, Matthew Tancik, and Pieter Abbeel.
\newblock Putting nerf on a diet: Semantically consistent few-shot view
  synthesis.
\newblock In {\em Proceedings of the IEEE/CVF International Conference on
  Computer Vision}, pages 5885--5894, 2021.

\bibitem{jang2021codenerf}
Wonbong Jang and Lourdes Agapito.
\newblock Codenerf: Disentangled neural radiance fields for object categories.
\newblock In {\em Proceedings of the IEEE/CVF International Conference on
  Computer Vision}, pages 12949--12958, 2021.

\bibitem{dtu}
Rasmus Jensen, Anders Dahl, George Vogiatzis, Engil Tola, and Henrik Aan{\ae}s.
\newblock {Large Scale Multi-view Stereopsis Evaluation}.
\newblock In {\em {2014 IEEE Conference on Computer Vision and Pattern
  Recognition}}, pages 406--413. IEEE, 2014.

\bibitem{jiang2021cotr}
Wei Jiang, Eduard Trulls, Jan Hosang, Andrea Tagliasacchi, and Kwang~Moo Yi.
\newblock Cotr: Correspondence transformer for matching across images.
\newblock In {\em Proceedings of the IEEE/CVF International Conference on
  Computer Vision}, pages 6207--6217, 2021.

\bibitem{conerf}
Kacper Kania, Kwang~Moo Yi, Marek Kowalski, Tomasz Trzci{\'n}ski, and Andrea
  Tagliasacchi.
\newblock Conerf: Controllable neural radiance fields.
\newblock In {\em Proceedings of the IEEE/CVF Conference on Computer Vision and
  Pattern Recognition}, pages 18623--18632, 2022.

\bibitem{stylegan}
Tero Karras, Samuli Laine, and Timo Aila.
\newblock A style-based generator architecture for generative adversarial
  networks.
\newblock In {\em Proceedings of the IEEE/CVF conference on computer vision and
  pattern recognition}, pages 4401--4410, 2019.

\bibitem{kasten2021layered}
Yoni Kasten, Dolev Ofri, Oliver Wang, and Tali Dekel.
\newblock Layered neural atlases for consistent video editing.
\newblock {\em ACM Transactions on Graphics (TOG)}, 40(6):1--12, 2021.

\bibitem{imagic}
Bahjat Kawar, Shiran Zada, Oran Lang, Omer Tov, Huiwen Chang, Tali Dekel, Inbar
  Mosseri, and Michal Irani.
\newblock Imagic: Text-based real image editing with diffusion models.
\newblock {\em arXiv preprint arXiv:2210.09276}, 2022.

\bibitem{kobayashi2022decomposing}
Sosuke Kobayashi, Eiichi Matsumoto, and Vincent Sitzmann.
\newblock Decomposing nerf for editing via feature field distillation.
\newblock {\em arXiv preprint arXiv:2205.15585}, 2022.

\bibitem{li2022vox}
Hai Li, Xingrui Yang, Hongjia Zhai, Yuqian Liu, Hujun Bao, and Guofeng Zhang.
\newblock Vox-surf: Voxel-based implicit surface representation.
\newblock {\em IEEE Transactions on Visualization and Computer Graphics}, 2022.

\bibitem{li2022symmnerf}
Xingyi Li, Chaoyi Hong, Yiran Wang, Zhiguo Cao, Ke Xian, and Guosheng Lin.
\newblock Symmnerf: Learning to explore symmetry prior for single-view view
  synthesis.
\newblock {\em arXiv preprint arXiv:2209.14819}, 2022.

\bibitem{neural_scene_flow}
Zhengqi Li, Simon Niklaus, Noah Snavely, and Oliver Wang.
\newblock Neural scene flow fields for space-time view synthesis of dynamic
  scenes.
\newblock In {\em Proceedings of the IEEE/CVF Conference on Computer Vision and
  Pattern Recognition}, pages 6498--6508, 2021.

\bibitem{editgan}
Huan Ling, Karsten Kreis, Daiqing Li, Seung~Wook Kim, Antonio Torralba, and
  Sanja Fidler.
\newblock Editgan: High-precision semantic image editing.
\newblock {\em Advances in Neural Information Processing Systems},
  34:16331--16345, 2021.

\bibitem{edit_nerf}
Steven Liu, Xiuming Zhang, Zhoutong Zhang, Richard Zhang, Jun-Yan Zhu, and
  Bryan Russell.
\newblock Editing conditional radiance fields.
\newblock In {\em Proceedings of the IEEE/CVF International Conference on
  Computer Vision}, pages 5773--5783, 2021.

\bibitem{lorensen1987marching}
William~E Lorensen and Harvey~E Cline.
\newblock {Marching Cubes: A High Resolution 3D Surface Construction
  Algorithm}.
\newblock {\em {ACM SIGGRAPH Computer Graphics}}, 21(4):163--169, 1987.

\bibitem{blenderswap}
John~Roper Matthew~Muldoon.
\newblock Blenderswap.
\newblock \url{https://www.blenderswap.com/}, 2022.
\newblock {A}ccessed: 2022-11-10.

\bibitem{quadrature_rule}
Nelson~L. Max.
\newblock {Optical Models for Direct Volume Rendering}.
\newblock {\em {IEEE} Trans. Vis. Comput. Graph.}, 1(2):99--108, 1995.

\bibitem{sdedit}
Chenlin Meng, Yang Song, Jiaming Song, Jiajun Wu, Jun-Yan Zhu, and Stefano
  Ermon.
\newblock Sdedit: Image synthesis and editing with stochastic differential
  equations.
\newblock {\em arXiv preprint arXiv:2108.01073}, 2021.

\bibitem{mi2022im2nerf}
Lu Mi, Abhijit Kundu, David Ross, Frank Dellaert, Noah Snavely, and Alireza
  Fathi.
\newblock im2nerf: Image to neural radiance field in the wild.
\newblock {\em arXiv preprint arXiv:2209.04061}, 2022.

\bibitem{mihajlovic2021leap}
Marko Mihajlovic, Yan Zhang, Michael~J Black, and Siyu Tang.
\newblock Leap: Learning articulated occupancy of people.
\newblock In {\em Proceedings of the IEEE/CVF Conference on Computer Vision and
  Pattern Recognition}, pages 10461--10471, 2021.

\bibitem{nerf}
Ben Mildenhall, Pratul~P Srinivasan, Matthew Tancik, Jonathan~T Barron, Ravi
  Ramamoorthi, and Ren Ng.
\newblock Nerf: Representing scenes as neural radiance fields for view
  synthesis.
\newblock {\em Communications of the ACM}, 65(1):99--106, 2021.

\bibitem{mirzaei2022laterf}
Ashkan Mirzaei, Yash Kant, Jonathan Kelly, and Igor Gilitschenski.
\newblock Laterf: Label and text driven object radiance fields.
\newblock {\em arXiv preprint arXiv:2207.01583}, 2022.

\bibitem{muller2022instant}
Thomas M{\"u}ller, Alex Evans, Christoph Schied, and Alexander Keller.
\newblock Instant neural graphics primitives with a multiresolution hash
  encoding.
\newblock {\em arXiv preprint arXiv:2201.05989}, 2022.

\bibitem{niemeyer2022regnerf}
Michael Niemeyer, Jonathan~T Barron, Ben Mildenhall, Mehdi~SM Sajjadi, Andreas
  Geiger, and Noha Radwan.
\newblock Regnerf: Regularizing neural radiance fields for view synthesis from
  sparse inputs.
\newblock In {\em Proceedings of the IEEE/CVF Conference on Computer Vision and
  Pattern Recognition}, pages 5480--5490, 2022.

\bibitem{niemeyer2021giraffe}
Michael Niemeyer and Andreas Geiger.
\newblock Giraffe: Representing scenes as compositional generative neural
  feature fields.
\newblock In {\em Proceedings of the IEEE/CVF Conference on Computer Vision and
  Pattern Recognition}, pages 11453--11464, 2021.

\bibitem{park2019deepsdf}
Jeong~Joon Park, Peter Florence, Julian Straub, Richard Newcombe, and Steven
  Lovegrove.
\newblock Deepsdf: Learning continuous signed distance functions for shape
  representation.
\newblock In {\em Proceedings of the IEEE/CVF conference on computer vision and
  pattern recognition}, pages 165--174, 2019.

\bibitem{photoshape2018}
Keunhong Park, Konstantinos Rematas, Ali Farhadi, and Steven~M. Seitz.
\newblock Photoshape: Photorealistic materials for large-scale shape
  collections.
\newblock {\em ACM Trans. Graph.}, 37(6), Nov. 2018.

\bibitem{park2021nerfies}
Keunhong Park, Utkarsh Sinha, Jonathan~T Barron, Sofien Bouaziz, Dan~B Goldman,
  Steven~M Seitz, and Ricardo Martin-Brualla.
\newblock Nerfies: Deformable neural radiance fields.
\newblock In {\em Proceedings of the IEEE/CVF International Conference on
  Computer Vision}, pages 5865--5874, 2021.

\bibitem{park2021hypernerf}
Keunhong Park, Utkarsh Sinha, Peter Hedman, Jonathan~T Barron, Sofien Bouaziz,
  Dan~B Goldman, Ricardo Martin-Brualla, and Steven~M Seitz.
\newblock Hypernerf: A higher-dimensional representation for topologically
  varying neural radiance fields.
\newblock {\em arXiv preprint arXiv:2106.13228}, 2021.

\bibitem{patashnik2021styleclip}
Or Patashnik, Zongze Wu, Eli Shechtman, Daniel Cohen-Or, and Dani Lischinski.
\newblock Styleclip: Text-driven manipulation of stylegan imagery.
\newblock In {\em Proceedings of the IEEE/CVF International Conference on
  Computer Vision}, pages 2085--2094, 2021.

\bibitem{poole2022dreamfusion}
Ben Poole, Ajay Jain, Jonathan~T Barron, and Ben Mildenhall.
\newblock Dreamfusion: Text-to-3d using 2d diffusion.
\newblock {\em arXiv preprint arXiv:2209.14988}, 2022.

\bibitem{clip}
Alec Radford, Jong~Wook Kim, Chris Hallacy, Aditya Ramesh, Gabriel Goh,
  Sandhini Agarwal, Girish Sastry, Amanda Askell, Pamela Mishkin, Jack Clark,
  et~al.
\newblock Learning transferable visual models from natural language
  supervision.
\newblock In {\em International Conference on Machine Learning}, pages
  8748--8763. PMLR, 2021.

\bibitem{diffusion}
Robin Rombach, Andreas Blattmann, Dominik Lorenz, Patrick Esser, and Björn
  Ommer.
\newblock High-resolution image synthesis with latent diffusion models, 2021.

\bibitem{CarWale}
Arun~Kumar Sahlam.
\newblock Carwale.
\newblock \url{https://www.carwale.com/}, 2022.
\newblock {A}ccessed: 2022-11-10.

\bibitem{sangkloy2017scribbler}
Patsorn Sangkloy, Jingwan Lu, Chen Fang, Fisher Yu, and James Hays.
\newblock Scribbler: Controlling deep image synthesis with sketch and color.
\newblock In {\em Proceedings of the IEEE conference on computer vision and
  pattern recognition}, pages 5400--5409, 2017.

\bibitem{schonberger2016structure}
Johannes~L Schonberger and Jan-Michael Frahm.
\newblock Structure-from-motion revisited.
\newblock In {\em Proceedings of the IEEE conference on computer vision and
  pattern recognition}, pages 4104--4113, 2016.

\bibitem{schwarz2020graf}
Katja Schwarz, Yiyi Liao, Michael Niemeyer, and Andreas Geiger.
\newblock Graf: Generative radiance fields for 3d-aware image synthesis.
\newblock {\em Advances in Neural Information Processing Systems},
  33:20154--20166, 2020.

\bibitem{as_rigid_as_possible}
Olga Sorkine and Marc Alexa.
\newblock As-rigid-as-possible surface modeling.
\newblock In {\em Symposium on Geometry processing}, volume~4, pages 109--116,
  2007.

\bibitem{Mixamo}
Nazim~Kareemi Stefano~Corazza.
\newblock Mixamo.
\newblock \url{https://www.mixamo.com/}, 2022.
\newblock {A}ccessed: 2022-11-10.

\bibitem{sun2022ide}
Jingxiang Sun, Xuan Wang, Yichun Shi, Lizhen Wang, Jue Wang, and Yebin Liu.
\newblock Ide-3d: Interactive disentangled editing for high-resolution 3d-aware
  portrait synthesis.
\newblock {\em arXiv preprint arXiv:2205.15517}, 2022.

\bibitem{sun2022fenerf}
Jingxiang Sun, Xuan Wang, Yong Zhang, Xiaoyu Li, Qi Zhang, Yebin Liu, and Jue
  Wang.
\newblock Fenerf: Face editing in neural radiance fields.
\newblock In {\em Proceedings of the IEEE/CVF Conference on Computer Vision and
  Pattern Recognition}, pages 7672--7682, 2022.

\bibitem{feature_fusion_field}
Vadim Tschernezki, Iro Laina, Diane Larlus, and Andrea Vedaldi.
\newblock Neural feature fusion fields: 3d distillation of self-supervised 2d
  image representations.
\newblock {\em arXiv preprint arXiv:2209.03494}, 2022.

\bibitem{splice}
Narek Tumanyan, Omer Bar-Tal, Shai Bagon, and Tali Dekel.
\newblock Splicing vit features for semantic appearance transfer.
\newblock In {\em Proceedings of the IEEE/CVF Conference on Computer Vision and
  Pattern Recognition}, pages 10748--10757, 2022.

\bibitem{vora2021nesf}
Suhani Vora, Noha Radwan, Klaus Greff, Henning Meyer, Kyle Genova, Mehdi~SM
  Sajjadi, Etienne Pot, Andrea Tagliasacchi, and Daniel Duckworth.
\newblock Nesf: Neural semantic fields for generalizable semantic segmentation
  of 3d scenes.
\newblock {\em arXiv preprint arXiv:2111.13260}, 2021.

\bibitem{clip_nerf}
Can Wang, Menglei Chai, Mingming He, Dongdong Chen, and Jing Liao.
\newblock Clip-nerf: Text-and-image driven manipulation of neural radiance
  fields.
\newblock In {\em Proceedings of the IEEE/CVF Conference on Computer Vision and
  Pattern Recognition}, pages 3835--3844, 2022.

\bibitem{neus}
Peng Wang, Lingjie Liu, Yuan Liu, Christian Theobalt, Taku Komura, and Wenping
  Wang.
\newblock Neus: Learning neural implicit surfaces by volume rendering for
  multi-view reconstruction.
\newblock {\em NeurIPS}, 2021.

\bibitem{wang2018high}
Ting-Chun Wang, Ming-Yu Liu, Jun-Yan Zhu, Andrew Tao, Jan Kautz, and Bryan
  Catanzaro.
\newblock High-resolution image synthesis and semantic manipulation with
  conditional gans.
\newblock In {\em Proceedings of the IEEE conference on computer vision and
  pattern recognition}, pages 8798--8807, 2018.

\bibitem{wu2022objectsdf}
Qianyi Wu, Xian Liu, Yuedong Chen, Kejie Li, Chuanxia Zheng, Jianfei Cai, and
  Jianmin Zheng.
\newblock Object-compositional neural implicit surfaces.
\newblock {\em arXiv preprint arXiv:2207.09686}, 2022.

\bibitem{xu2022sinnerf}
Dejia Xu, Yifan Jiang, Peihao Wang, Zhiwen Fan, Humphrey Shi, and Zhangyang
  Wang.
\newblock Sinnerf: Training neural radiance fields on complex scenes from a
  single image.
\newblock {\em arXiv preprint arXiv:2204.00928}, 2022.

\bibitem{neumesh}
Bangbang Yang, Chong Bao, Junyi Zeng, Hujun Bao, Yinda Zhang, Zhaopeng Cui, and
  Guofeng Zhang.
\newblock Neumesh: Learning disentangled neural mesh-based implicit field for
  geometry and texture editing.
\newblock In {\em European Conference on Computer Vision}, pages 597--614.
  Springer, 2022.

\bibitem{yang2022neural}
Bangbang Yang, Yinda Zhang, Yijin Li, Zhaopeng Cui, Sean Fanello, Hujun Bao,
  and Guofeng Zhang.
\newblock Neural rendering in a room: amodal 3d understanding and
  free-viewpoint rendering for the closed scene composed of pre-captured
  objects.
\newblock {\em ACM Transactions on Graphics (TOG)}, 41(4):1--10, 2022.

\bibitem{object_nerf}
Bangbang Yang, Yinda Zhang, Yinghao Xu, Yijin Li, Han Zhou, Hujun Bao, Guofeng
  Zhang, and Zhaopeng Cui.
\newblock Learning object-compositional neural radiance field for editable
  scene rendering.
\newblock In {\em Proceedings of the IEEE/CVF International Conference on
  Computer Vision}, pages 13779--13788, 2021.

\bibitem{yang2022vox}
Xingrui Yang, Hai Li, Hongjia Zhai, Yuhang Ming, Yuqian Liu, and Guofeng Zhang.
\newblock Vox-fusion: Dense tracking and mapping with voxel-based neural
  implicit representation.
\newblock In {\em 2022 IEEE International Symposium on Mixed and Augmented
  Reality (ISMAR)}, pages 499--507. IEEE, 2022.

\bibitem{volsdf}
Lior Yariv, Jiatao Gu, Yoni Kasten, and Yaron Lipman.
\newblock Volume rendering of neural implicit surfaces.
\newblock {\em arXiv preprint arXiv:2106.12052}, 2021.

\bibitem{nerf_editing}
Yu-Jie Yuan, Yang-Tian Sun, Yu-Kun Lai, Yuewen Ma, Rongfei Jia, and Lin Gao.
\newblock Nerf-editing: geometry editing of neural radiance fields.
\newblock In {\em Proceedings of the IEEE/CVF Conference on Computer Vision and
  Pattern Recognition}, pages 18353--18364, 2022.

\bibitem{arf}
Kai Zhang, Nick Kolkin, Sai Bi, Fujun Luan, Zexiang Xu, Eli Shechtman, and Noah
  Snavely.
\newblock Arf: Artistic radiance fields.
\newblock In {\em European Conference on Computer Vision}, pages 717--733.
  Springer, 2022.

\bibitem{zhang2021nerfactor}
Xiuming Zhang, Pratul~P Srinivasan, Boyang Deng, Paul Debevec, William~T
  Freeman, and Jonathan~T Barron.
\newblock Nerfactor: Neural factorization of shape and reflectance under an
  unknown illumination.
\newblock {\em ACM Transactions on Graphics (TOG)}, 40(6):1--18, 2021.

\bibitem{zhu2022nice}
Zihan Zhu, Songyou Peng, Viktor Larsson, Weiwei Xu, Hujun Bao, Zhaopeng Cui,
  Martin~R Oswald, and Marc Pollefeys.
\newblock Nice-slam: Neural implicit scalable encoding for slam.
\newblock In {\em Proceedings of the IEEE/CVF Conference on Computer Vision and
  Pattern Recognition}, pages 12786--12796, 2022.

\end{thebibliography}
}
\clearpage

\appendix

\renewcommand\thesection{\Alph{section}}
\renewcommand\thetable{\Alph{table}}
\renewcommand\thefigure{\Alph{figure}}

\begin{strip}
\begin{center}
{\huge \bf Supplementary Material}
\end{center}
\end{strip}

\vspace{0.5em}

\renewcommand\thesection{\Alph{section}}
\renewcommand\thetable{\Alph{table}}
\renewcommand\thefigure{\Alph{figure}}

In this supplementary material, we describe more details of our method, including model architecture in Sec.~\ref{sec:model_arch}, dataset preparation in Sec.~\ref{sec:dataset_prepare}, and implementation details in Sec.~\ref{sec:impl}.
Besides, we also conduct more experiments in Sec.~\ref{sec:more_expr}.
More qualitative results can be found in our supplementary video, and the source code will be released upon the acceptance of this paper.

\section{Model Architecture}
\label{sec:model_arch}

We first explain the details of the model architecture.
Specifically, we adopt the multi-resolution voxel-hashing encoder by Müller \etal~\cite{muller2022instant} as the coordinate-based encoder, and build the template NeRF and the editing field in a decoupled manner.
The voxel-hashing encoder is constructed with 16 levels with 2-dimensional features for each level.
For the template NeRF, we use the voxel-hashing encoder to encode the queries' coordinates and use spherical harmonics with 4 degrees to encode the ray direction.
The density and color heads for model output consist of 1 hidden layer with 128 hidden size and 2 hidden layers with 64 hidden size, respectively.
As introduced in Sec.~\textcolor{red}{3.1}, the editing field consists of a geometric modification field $F_{\Delta G}$ and a texture modification field $F_{\Delta T}$.
The geometric modification field $F_{\Delta G}$ and the corresponding forward modification field $F_{\Delta G}'$ are both constructed with an MLP of 1 hidden layer and 128 hidden size with the ReLU activation, and we adopt the positional encoding~\cite{nerf} (with 4 frequencies) to all input query points.
The texture modification field $F_{\Delta T}$ is constructed with a voxel-hashing encoder (same size as the template NeRF), followed by an MLP of 1 hidden layer and 128 hidden size with ReLU activation.
During the dual volume rendering stage, we follow Mildenhall \etal~\cite{nerf} by using 64 coarse samples and 128 fine samples for each ray, and render the deformed template image $\hat{I}_o$ and color modification image $\hat{I}_{m}$ with the same density values.
Then, as explained in Sec.~\textcolor{red}{3.3}, we use a color compositing layer to obtain the edited view $\hat{I}$ by blending $\hat{I}_{m}$ into $\hat{I}_o$, where the color compositing layer is constructed using a compact UNet-like structure (with 2-layer encoder (3 $\rightarrow$ 16 $\rightarrow$ 32) and a symmetrical decoder, all layers comprise 3$\times$3 convolutions).
Besides, we can integrate temporal attribute~\cite{park2021nerfies,park2021hypernerf} (from 0 to 1) to the input of $F_{\Delta G}$ (with the positional encoding of 4 frequencies), and train the geometric editing on the edited transitions with temporal attributes as conditions, \eg, the dynamic motion effect shown in the supplementary video.

\begin{figure}[!t]
    \centering
    \includegraphics[width=1.0\linewidth, trim={0 0 0 0}, clip]{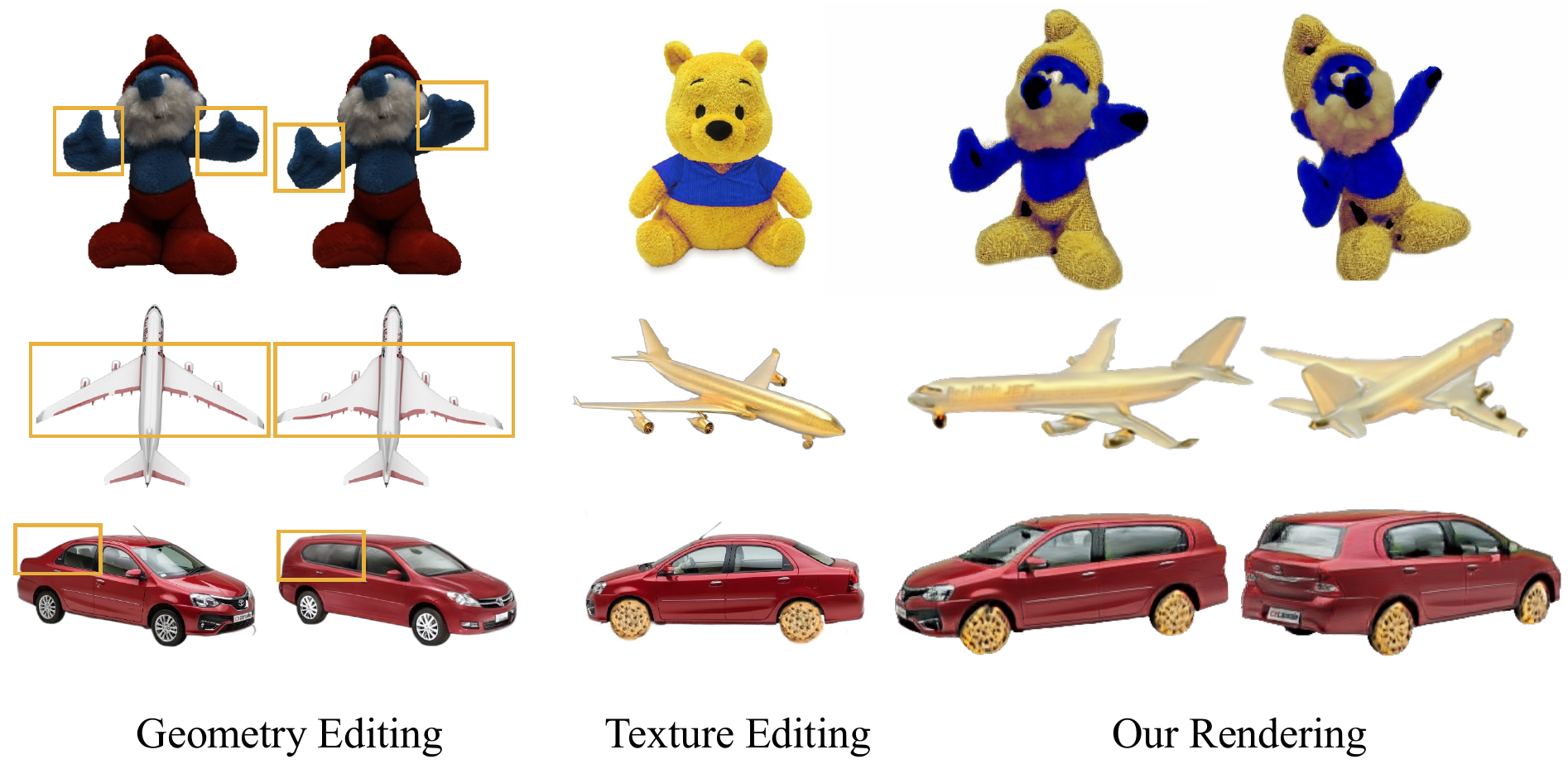}
    \caption{
    We show examples of hybrid object editing by combining geometric and texture editing.
}
    \label{fig:supp_hybrid_editing}
\end{figure}
\section{Dataset Preparation}
\label{sec:dataset_prepare}

We evaluate SINE on both real-world/synthetic and object/scene datasets.
Specifically, for the real-world car datasets~\cite{CarWale}, each sequence contains 72 images with a car rotating on the turntable.
We use Colmap~\cite{schonberger2016structure} to recover camera poses w.r.t the cars' centers for all the images.
For the data of Photoshape~\cite{photoshape2018}, EditNeRF~\cite{edit_nerf} does not provide edited GT images,  so we regenerate all testing cases using Blender, which is more challenging than the original ones (\eg, we stretch the whole chair or enlarge holes, while EditNeRF~\cite{edit_nerf} only fills a tiny hole or removes legs).
For the data Blenderswap~\cite{blenderswap}, we render the scenes with Blender’s Cycle engine with realistic environment HDR maps. 
For users' 2D image editing, we use Adobe After Effect / Photoshop to deform images (geometric editing) and paint patterns (texture editing), and use EditGAN~\cite{editgan} and Text2LIVE~\cite{text2live} to edit images with semantic strokes or text-prompts.
For users' target images (\eg, cars and chairs) from the Internet, we remove their backgrounds using remove.bg~\cite{removebg} before conducting texture editing.

\begin{figure*}[!t]
    \centering
    \includegraphics[width=1.0\linewidth, trim={0 0 0 0}, clip]{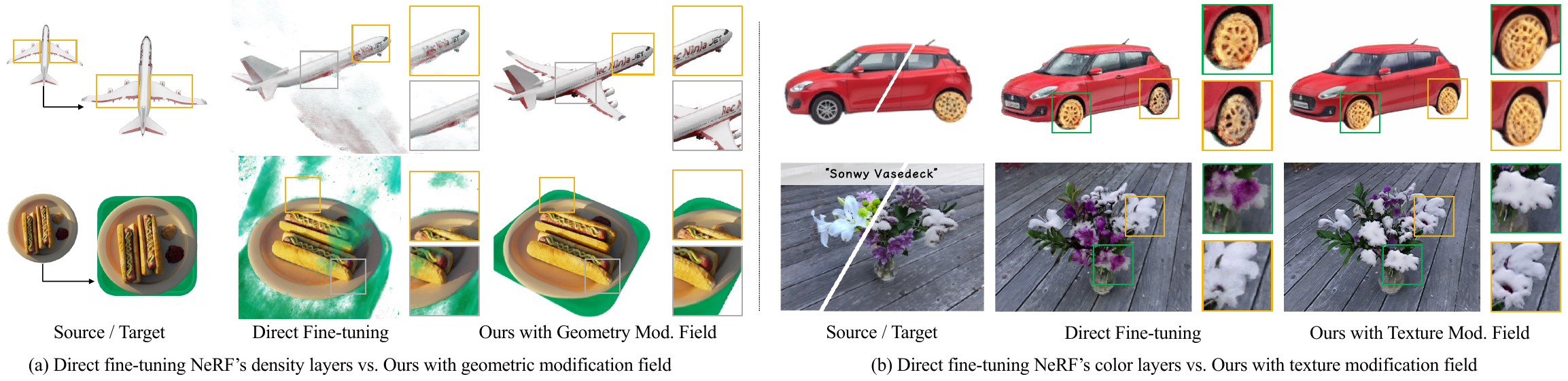}
    \caption{
    We compare our editing field with directly fine-tuning template NeRF.
}
    \label{fig:supp_direct_finetune}
\end{figure*}

\begin{figure*}[!t]
    \centering
    \includegraphics[width=1.0\linewidth, trim={0 0 0 0}, clip]{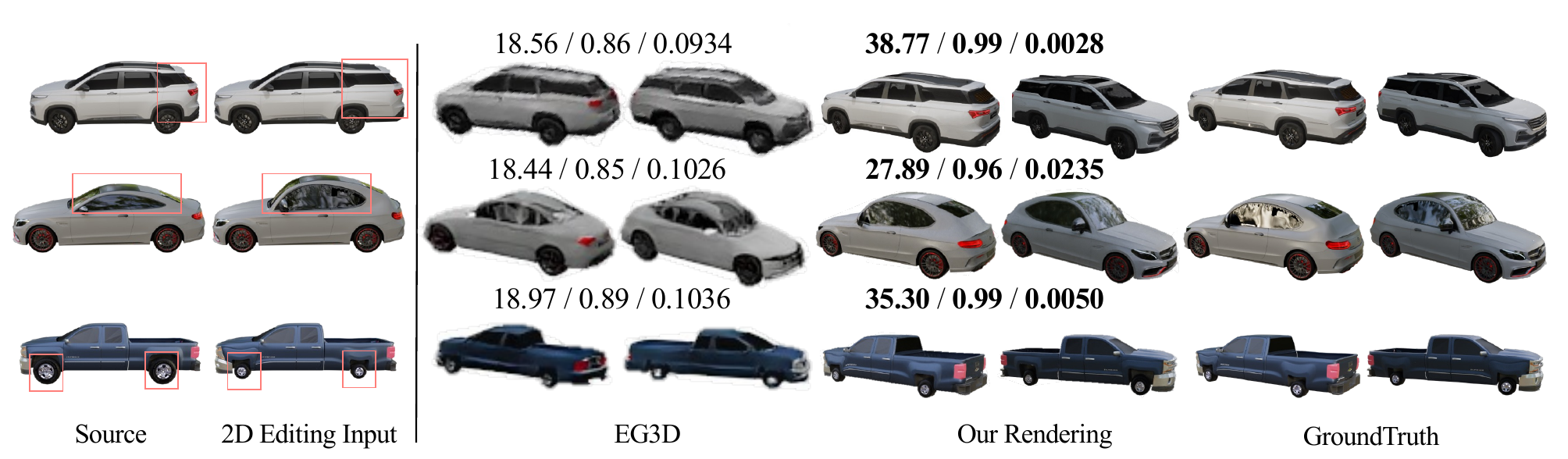}
    \caption{
    We show the quantitative comparison between our method and EG3D~\cite{eg3d} on the synthetic cars~\cite{blenderswap}, where the metrics of PSNR$\uparrow$ / SSIM$\uparrow$ / LPIPS$\downarrow$ are annotated above.
}
    \label{fig:supp_geo_syncar}
\end{figure*}

\section{Implementation Details}
\label{sec:impl}

\noindent\textbf{Training details.}
As introduced in the main paper, our method performs semantic-driven editing upon the given NeRF model.
Specifically, for each object or scene, we first train a generic template NeRF model.
Then, we learn geometric editing and texture editing with editing from a single perspective.
For geometric editing, since the geometric changes are sometimes combined with minor color changes, we also fine-tune the color modification field with a photometric loss (see the first term in Eq.~(\textcolor{red}{3})).
For texture editing, the texture transferring loss (Eq.(\textcolor{red}{6}))is defined on a complete image, which is not compatible with NeRF's sparse ray supervision.
Therefore, we adopt the deferred back-propagation technique from Zhang \etal~\cite{arf} for texture editing.
Practically, we first render the full-sized $\hat{I}_o$ and $\hat{I}_{m}$ and forward the color compositing layer, and then compute the losses to cache the complete image gradient w.r.t the $\hat{I}_m$ and the color compositing layer, and re-render the $\hat{I}$ and back-propagate the gradients to the $F_{\Delta T}$ and the color compositing layer in a patch-wise manner.
To make a smooth convergence, we take the coarse-to-fine regularization~\cite{park2021nerfies} on the color modification field by progressively increasing the frequency band of the input features during the training process.
Furthermore, we randomly perturb the pose to augment the data distribution and avoid the overfitting of the texture editing.
The whole training process of the template NeRF and our editing field takes about 12 hours on a single Nvidia RTX 3090 graphics card.

\noindent\textbf{Preparation of proxy mesh in geometric editing.}
As introduced in Sec.~\textcolor{red}{3.2}, we use a proxy mesh to represent NeRF's geometry during geometric editing.
In practice, we directly obtain the proxy mesh using off-the-shelf tools (\ie, implicit surface reconstruction method NeuS~\cite{neus}).
Since we optimize DIF~\cite{dif} latent code $\hat{\textbf{z}}$ and deform the proxy mesh $\hat{M}$ during the editing, the initial proxy mesh should be binding to a latent code beforehand.
Therefore, we obtain the initial latent code $\hat{\textbf{z}}$ to the corresponding initial proxy mesh $\hat{M}_\sigma$ in an auto-decoding manner~\cite{park2019deepsdf,dif} before training.

\noindent\textbf{Cycle loss in geometric editing.}
During the training of geometric editing, we additionally train a forward modification field $F_{\Delta G}'$ to map the template proxy mesh to the edited space.
The forward modification field $F_{\Delta G}'$ and the implicit geometric modification field $F_{\Delta G}$ are both supervised with an cycle loss~\cite{mihajlovic2021leap,neural_scene_flow}, which is defined as:
\begin{equation}
\begin{split}
\mathcal{L}_{\text{cycle}} = \frac{1}{M}\sum_{i =1}^M & ||F_{\Delta G} (F_{\Delta G}'(\mathbf{p}_i)) - \mathbf{p}_i|| + \\ 
&||F_{\Delta G}' (F_{\Delta G}(\mathbf{p}_i)) - \mathbf{p}_i||,
\end{split}
\end{equation}
where $\{\mathbf{p}_i|i=1,...,M\}$ are the uniform point samples in 3D space, and we set $M=1000$ in our experiment.

\noindent\textbf{Feature-cluster-based semantic masking.}
As introduced in Sec.~\textcolor{red}{3.4}, we train a 3D feature field with DINO-ViT's feature maps, and generate feature clusters from the user-painted regions, which will be used to compute semantic masks to distinguish foreground editing areas and background areas.
Specifically, we first render the feature map under the specific editing view, and sample 1000 feature points on the user's painted region (which is directly accessible from the editing tools).
Then, we use K-Means to generate $K=15$ clusters from the sampled feature points.
During the training stage, we first render the current training view's feature map, and compute the L2-normalized pixel-wise feature distance (from 0 to 1) to the nearest clusters.
The pixels with distances smaller than 0.5 would be marked as foreground objects, and the others would be marked as background.
These computed editing masks would be used to regularize both geometric and texture editing (see Eq.~(\textcolor{red}{7})) to maintain the irrelevant content unchanged.

\noindent\textbf{User study.}
The questionnaire contains 17 cases, 8 for target-image-based editing (\eg, Fig.~\textcolor{red}{7} (a)) and 9 for text-prompts editing (\eg, Fig.~\textcolor{red}{7} (b)).
We show the participants a source image, a target image/text prompts, as well as the results produced by different methods. 
Participants are asked to select one result that best matches the style of the target image or the text meaning.

\begin{figure}[!t]
    \centering
    \includegraphics[width=1.0\linewidth, trim={0 0 0 0}, clip]{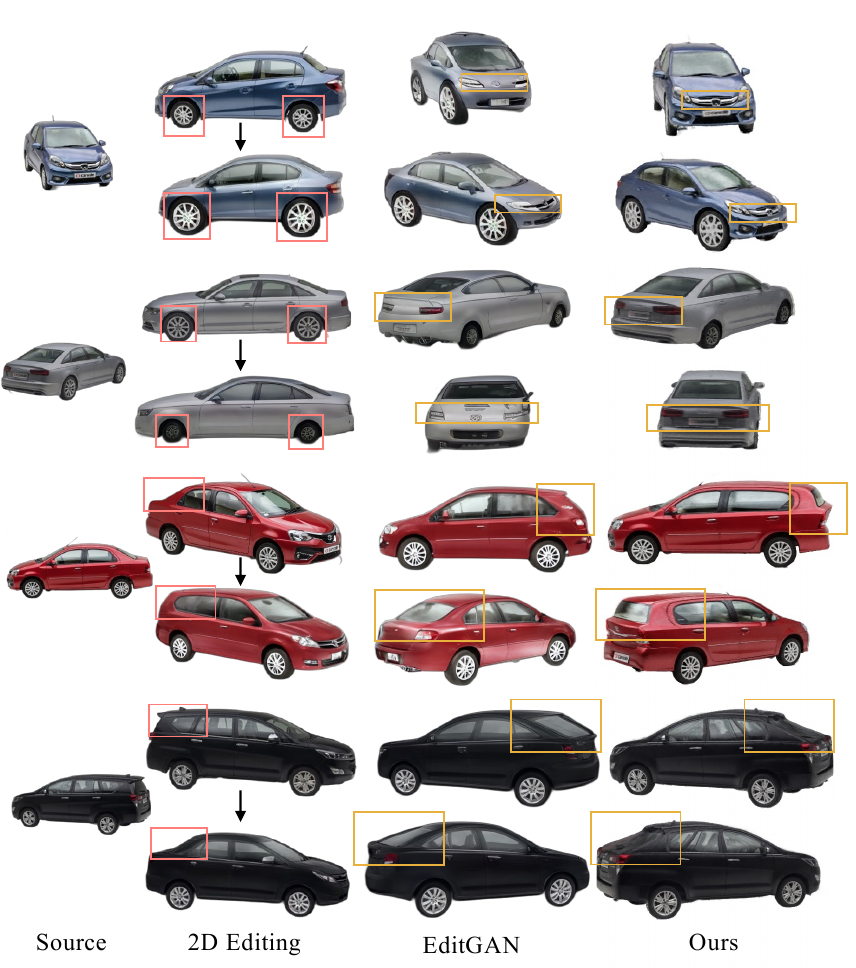}
    \caption{
    We show the comparison of our method with EditGAN~\cite{editgan} on the real-world car dataset~\cite{CarWale}.
}
    \label{fig:supp_geo_2dgan}
    \vspace{1.0em}

\end{figure}

\begin{figure}[!t]
    \centering
    \includegraphics[width=1.0\linewidth, trim={0 0 0 0}, clip]{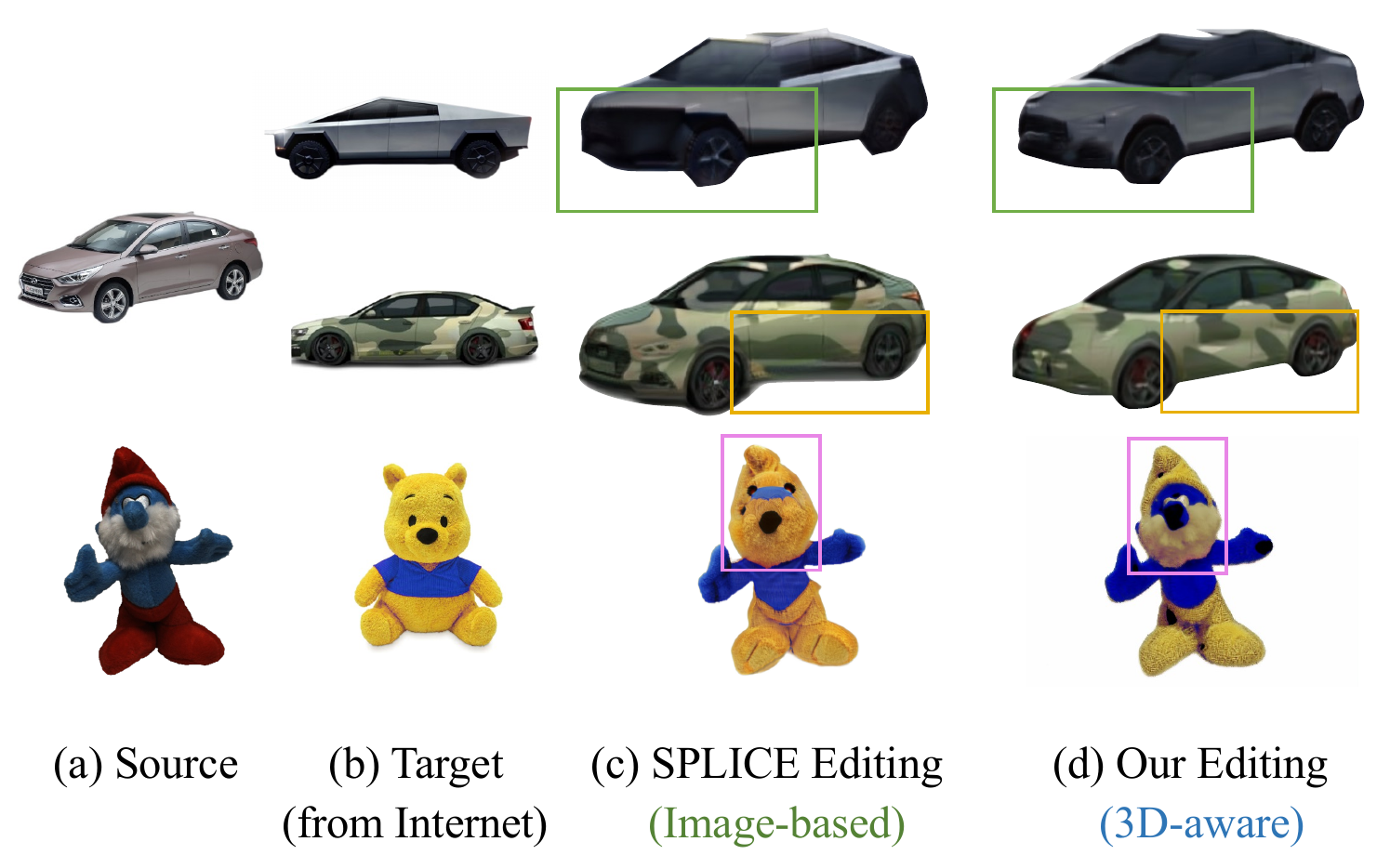}
    \caption{
    We compare our method with SPLICE-ViT~\cite{splice} on the real-world cars~\cite{CarWale} and toys~\cite{dtu}.
    Since our editing method is built upon 3D-aware models, we consistently achieve better texture-transferring results than SPLICE-ViT when the source and target are observed from different perspectives of views (\eg, cars) or with significant different shapes (\eg, plush toys).
}
    \label{fig:compare_2d_splice}
    \vspace{1.0em}
\end{figure}

\section{More Experiments}
\label{sec:more_expr}

\noindent\textbf{Hybrid editing with geometric and texture changes.}
We can combine geometric and texture editing on the same object by optimizing geometric-related losses and texture-transferring losses in turns.
As shown in Fig.~\ref{fig:supp_hybrid_editing}, we can edit objects' geometries while transferring textures with users' target images, \eg, the plush toy raises its hands and is painted in new textures from a yellow bear, and the airplane extends its wings and is painted golden.
Please refer to our supplementary video for a vivid animation of these effects.

\noindent\textbf{3D editing field vs. template NeRF fine-tuning.}
In this experiment, we compare our 3D editing field with the na\"ive fine-tuning template NeRF (which is adopted by CLIP-NeRF~\cite{clip_nerf} and DFF~\cite{kobayashi2022decomposing}).
Editing NeRF with only a single image is fairly ambiguous without external supervision (\eg, semantic hints). 
For a fair comparison, we provide external supervision for the baseline method (vanilla NeRF).
Specifically, for texture editing, we enable NeRF's related network layers to be optimized and use the same texture editing losses.
Directly fine-tuning NeRF's color layers can change the objects' texture to some extent but cannot reach the same quality as our full model (\eg, uncovered cookie tires and snowy flowers in Fig.~\ref{fig:supp_direct_finetune} (b)).
For geometry editing, we fine-tune vanilla NeRF with $\mathcal{L}_{\text{gt}}$ 
and $\mathcal{L}_{\text{reg}}$ 
since it is not trivial to apply SDF-based shape priors to vanilla NeRF.
As demonstrated in Fig.~\ref{fig:supp_direct_finetune} (a), na\"ively fine-tuning NeRF on geometric editing would lead to the overfitting to a single view, and the multi-view consistency is no longer ensured (\eg, broken wings and green floaters in Fig.~\ref{fig:supp_direct_finetune} (a)).

\noindent\textbf{Quantitative comparison with EG3D on synthetic cars.}
We conduct quantitative comparisons with the SOTA 3D-aware GAN method EG3D \cite{eg3d} on the synthetic car dataset. 
To obtain the ground-truth images of the edited results, we use Blender to render the training and testing views, and modify the cars' geometry within the software.
As shown in Fig~\ref{fig:supp_geo_syncar}, our method achieves better rendering quality than EG3D on both visual quality and all the metrics (PSNR, SSIM, and LPIPS).
For example, we can preserve the specular effect even after the editing (\eg, the specular area facing the light source and the reflection on the windshield), while EG3D struggles to produce photo-realistic results due to the limitation of its learned 3D latent representation.

\noindent\textbf{Comparison with 2D GANs.}
We compare our method against the SOTA 2D semantic editing method EditGAN\cite{editgan} on the real-world car dataset~\cite{CarWale}.
To make a fair comparison, we train our NeRF's backbone and EditGAN's style codes on all the multi-view images (\ie, each style code corresponds to one view).
Then, we perform semantic 2D editing on one single view using EditGAN.
For our method, we use the edited view to train our editing field.
And for EditGAN, we save the intermediate editing vector and add the editing vector to all the style codes, which yields multi-view edited images.
As demonstrated in Fig~\ref{fig:supp_geo_2dgan}, since EditGAN is agnostic to the 3D geometry, its results suffer from the inconsistent issue between different views, \eg, poor inversion results for the head and tail of the car, and the semantic editing result cannot be precisely applied to all views.
In contrast, our methods can synthesize cars with multi-view consistency and high-quality editing results.

\begin{figure}[!t]
    \centering
    \includegraphics[width=0.95\linewidth, trim={0 0 0 0}, clip]{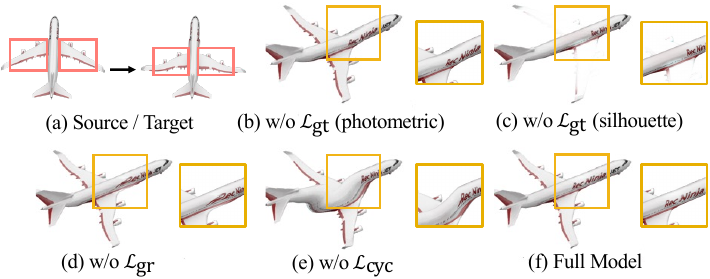}
    \caption{
    We inspect the efficacy of different constraints in geometric editing.
    }
    \label{fig:ablation_geo_loss}
        \vspace{1.0em}

\end{figure}

\noindent\textbf{More ablation studies on geometric supervision.} As shown in Fig.~\ref{fig:ablation_geo_loss}, since ablating $\mathcal{L}_{\text{gt}}$ (Eq.~(\textcolor{red}{3})) makes it no supervision on editing, we split it into photometric (b) and silhouette (c) terms, and the absence of either will result in distorted or washed-out texture.
\textbf{(d)}: When ablating deformation reg. loss $\mathcal{L}_{\text{gr}}$ (Eq.~(\textcolor{red}{4})), the edited object is severely distorted (\eg, the letters are stretched).
\textbf{(e)}: The cycle loss $\mathcal{L}_{\text{cyc}}$ (Eq.~(\textcolor{red}{1}) in supp.) brings constraints from shape prior to geometric mod. field $F_{\Delta G}$, and ablating it would lose the efficacy of semantic guidance (\eg, the twisted airplane).

\begin{figure}[!t]
    \centering
    \includegraphics[width=0.95\linewidth, trim={0 0 0 0}, clip]{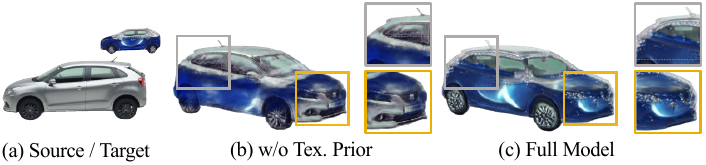}
    \caption{
    We inspect the efficacy of the texture prior constraint in texture editing
    }
    \label{fig:ablation_tex_loss}
    \vspace{1.0em}

\end{figure}

\noindent\textbf{Ablation study of texture supervision.}
In Fig.~\ref{fig:ablation_tex_loss}, we disable texture transfer loss $\mathcal{L}_\text{tex}$ (Eq.~(\textcolor{red}{6})) and utilize photometric loss to paint the target texture,
which leads to incomplete texture transferring results for invisible parts as shown below.
Besides, without texture prior supervision, we cannot transfer textures between objects with different shapes.

\noindent\textbf{Comparison of texture editing with image-based SPLICE-ViT.}
Our texture transferring loss (Eq.~(\textcolor{red}{6})) is inspired from SPLICE-ViT~\cite{splice}, but fully leverages the multi-view training scheme.
Therefore, we compare our texture editing with image-based SPLICE-ViT in Fig.~\ref{fig:compare_2d_splice}.
As shown in Fig.~\ref{fig:compare_2d_splice}, SPLICE-ViT is sensitive to the perspective difference of the source and target images, which results in overfitting appearances on the edited view, \eg, horizontal straight patterns of cars when observing cars from a slightly tilted view, distorted faces of the plush toy.
By contrast, our method consistently achieves better texture-transferring results with color patterns properly aligned to the cars' geometries and the plush toy's body parts.

\begin{figure}[!t]
    \centering
    \includegraphics[width=1.0\linewidth, trim={0 0 0 0}, clip]{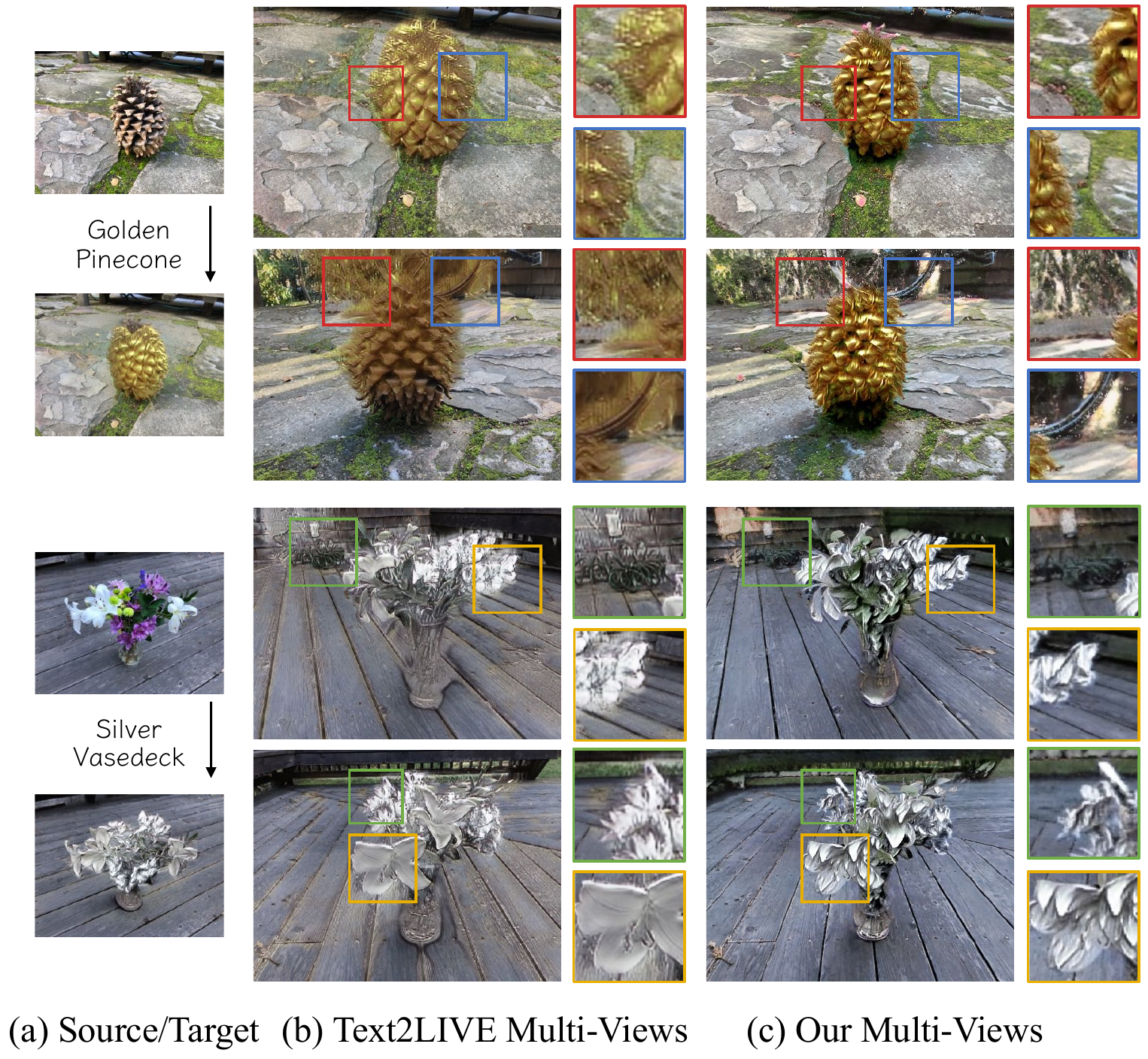}
    \caption{
    We compare our method with Text2LIVE on texture editing, where our method achieves better multi-view consistency. See the text for details.
}
    \label{fig:compare_text2live}
        \vspace{-1.0em}

\end{figure}

\noindent\textbf{Comparison of texture editing with Text2LIVE.}
As shown in Sec.~\textcolor{red}{4.4} from the main paper, since our method only requires one single-view image as editing input, we can naturally achieve text-prompt-based texture editing by cooperating with off-the-shelf text-driven editing methods (such as Text2LIVE~\cite{text2live}).
A follow-up question is, how does the Text2LIVE itself perform to the same 360$^{\circ}$ dataset in our texture editing task?
For video editing, Text2LIVE uses layered atlas~\cite{kasten2021layered} to convert objects and backgrounds into separated 2D layers.
However, in the unbounded 360$^{\circ}$ dataset (\eg, pinecone and vasedeck~\cite{nerf}), there is no proper way to unwrap 3D objects and scenes into 2D layers (and we also failed to train layered atlas on these 360$^{\circ}$ datasets).
Therefore, we directly apply its converged editing generator to the multi-view images.
As shown in Fig.~\ref{fig:compare_text2live}, although Text2LIVE produces similar-looking edited images, it cannot maintain multi-view consistency when the viewpoint changes (\eg, blurry edges at the golden pinecone, uncovered petals at the silver vasedeck, and the occasionally affected background).
On the contrary, our method naturally takes advantage of multi-view training and consistently delivers more plausible and realistic novel views.

\begin{figure}[!t]
    \centering
    \includegraphics[width=1.0\linewidth, trim={0 0 0 0}, clip]{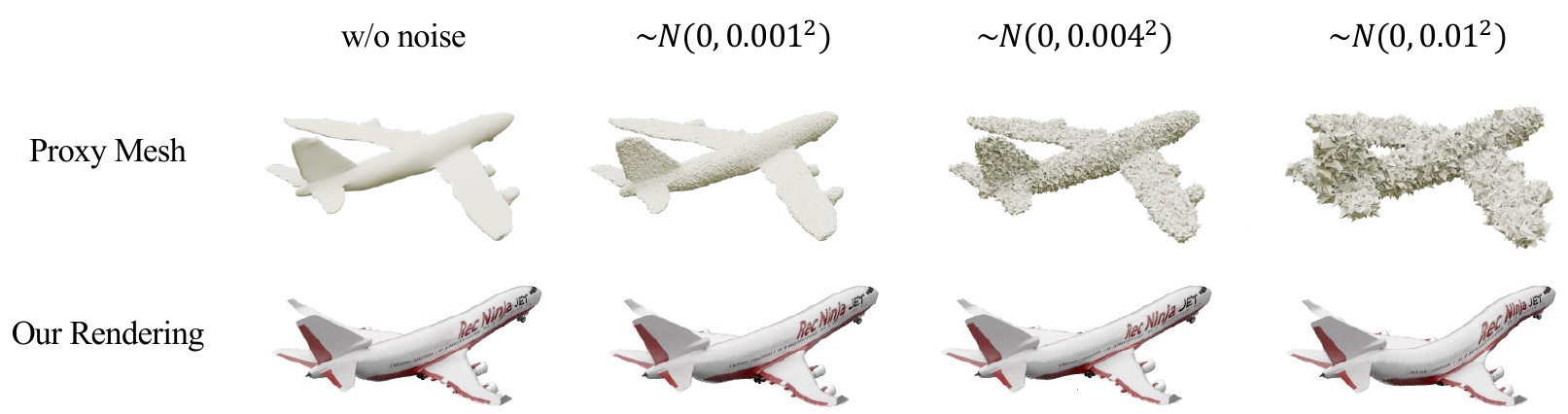}
    \caption{
    We show the robustness of geometric editing on proxy meshes with different qualities.
    The proxy meshes are jittered by adding gaussian noise with different variances (from $0.001^2$ to $0.01^2$).
}
    \label{fig:supp_geo_meshquality}
    \vspace{1.0em}
\end{figure}

\begin{figure*}[!t]
    \centering
    \includegraphics[width=1.0\linewidth, trim={0 0 0 0}, clip]{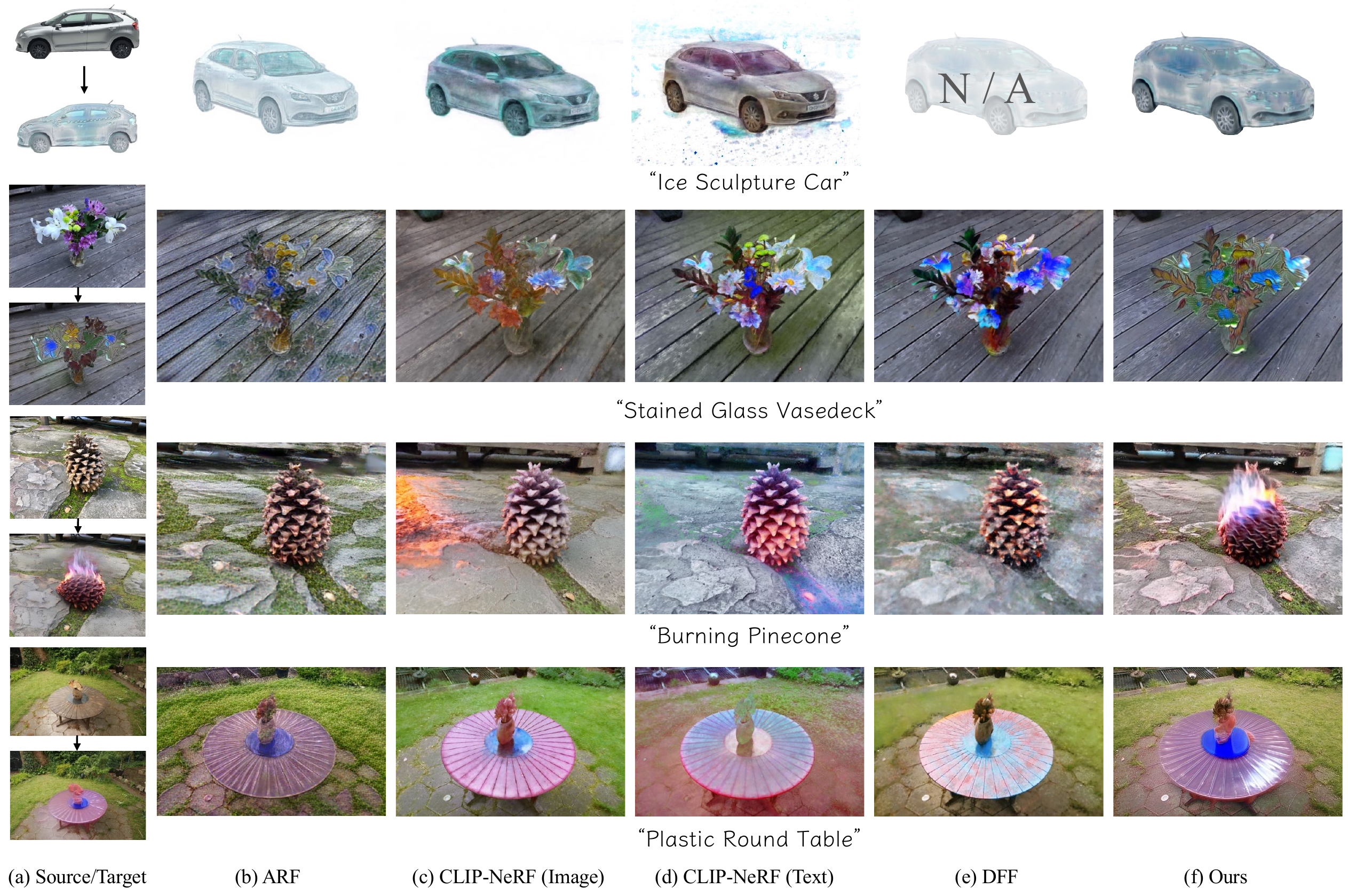}
    \caption{
    We show more comparison results of texture editing with ARF, CLIP-NeRF, and DFF on the real-world car~\cite{CarWale} and 360$^{\circ}$ scene dataset~\cite{nerf}.
    Our method consistently achieves more realistic and appealing editing results than the others.
}
    \label{fig:tex_edit_compare}
\end{figure*}

\begin{figure}[!t]
    \centering
    \includegraphics[width=0.95\linewidth, trim={0 0 0 0}, clip]{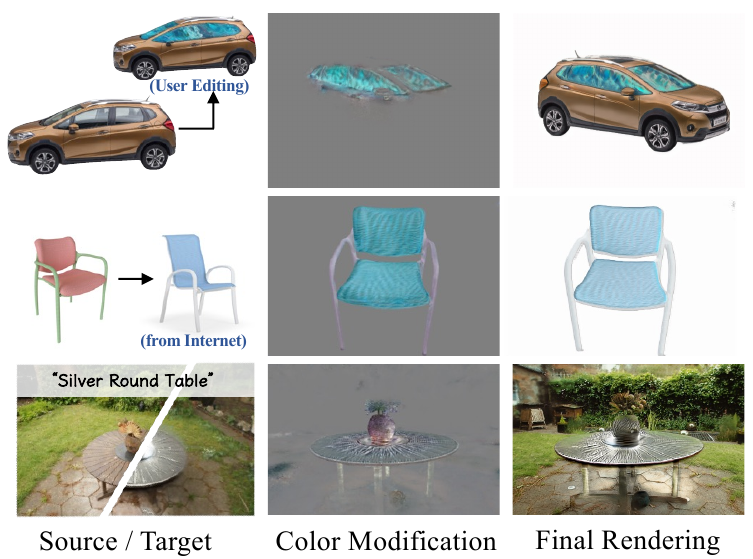}
    \caption{
    We show more rendering results from color modification field and compositional layer.
    }
    \label{fig:mod_rgb}
\end{figure}

\noindent\textbf{Robustness to the noise of proxy mesh.}
The geometry prior guidance uses the proxy mesh to supervise the geometry modification field.
Therefore, we analyze the effect of mesh quality on our editing results.
Specifically, we add 3 groups of gaussian noise to the vertices of the proxy mesh and conduct training of our editing field.
As shown in Fig~\ref{fig:supp_geo_meshquality}, our method can robustly learn geometric editing even with noisy proxy mesh (\eg, with the gaussian noise of $N(0, 0.004^2)$ in the third column).

\noindent\textbf{More comparison results on texture editing.}
We show more comparison results on the texture editing task with ARF~\cite{arf}, CLIP-NeRF~\cite{clip_nerf} and DFF~\cite{kobayashi2022decomposing} in Fig.~\ref{fig:tex_edit_compare} (where Fig.~\ref{fig:tex_edit_compare} (a) is the source view and the target view produced by Text2LIVE~\cite{text2live}).
For CLIP-NeRF~\cite{clip_nerf}, since the official codebase has not been fully released, we use our own implementation by fine-tuning NeRF's color-related field with CLIP loss, and both use the target features from text embedding and the image embedding (with the same target images in Fig.~\ref{fig:tex_edit_compare} (a)), which are denoted as CLIP-NeRF (Image) and CLIP-NeRF (Text), respectively.
For DFF~\cite{kobayashi2022decomposing}, we adopt the official codebase and use the texts for NeRF editing and background ray-filtering according to the document.
In Fig.~\ref{fig:tex_edit_compare}, we omit the DFF's object-centric comparison on the car, since it mainly focuses on scene-level decomposition and editing.
As demonstrated in Fig.~\ref{fig:tex_edit_compare}, NeRF stylization methods like ARF cannot precisely edit fine-grained effects on the desired location.
NeRF fine-tuning approaches like CLIP-NeRF and DFF only change appearance colors, but cannot produce vivid effects (\eg, the burning pinecone or ice sculpture cars).
Note that although DFF uses the semantic-field guided decomposed rendering to maintain the background color unchanged, this strategy is not compatible with our color compositing mechanism since we introduce an additional 2D CNN layer to blend the template and editing color for better visual appearance.

By contrast, 
our method both achieves realistic and appealing editing effects, and also effectively preserves background content, and the results are consistently preferred by most of the participants in the user study (see Sec.~\textcolor{red}{4.4}).

\noindent\textbf{Impact of texture modification field \& color compositional layer.}
The texture modification field learns detailed modifications and the compositional layer blends the original and modified rendering to produce the final edited results, as demonstrated in Sec. \textcolor{red}{4.5} and Fig. \textcolor{red}{8} (a).
Here we show more rendered texture mod. field (a.k.a. color mod. $\hat{I}_m$) in Fig~\ref{fig:mod_rgb}.

\begin{figure}[!t]
    \centering
    \vspace{0.0em}
    \includegraphics[width=0.95\linewidth, trim={0 0 0 0}, clip]{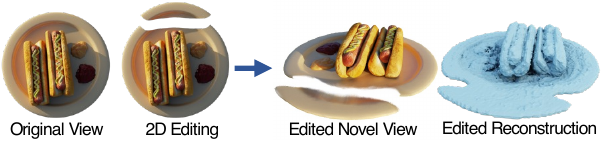}
    \caption{
    We show the results of geometry editing with topology changes.
    }
    \label{fig:hotdog_splitplate}
    \vspace{1.0em}
\end{figure}

\noindent\textbf{Deformation with topology changes.}
Our method does not support deformation with topology changes such as breaking the plate, but can provide a visually plausible result by making the ``broken part'' white, as shown in Fig~\ref{fig:hotdog_splitplate}.
In the future, we can integrate more flexible representations such as ambient slicing surface~\cite{park2021hypernerf} into our model.

\end{document}